\documentclass{article}

\usepackage{arxiv}
\usepackage{amssymb}
\usepackage{algorithm}
\usepackage{chngcntr}
\usepackage{algpseudocode}
\usepackage{url} 
\usepackage{booktabs}
\usepackage{subcaption}
\usepackage{graphicx}    
\usepackage{epsfig}
\usepackage{amsmath}
\usepackage{lineno}
\usepackage{lscape}
\usepackage{hyperref} 
\title{Unsupervised Port Berth Identification from Automatic Identification System Data}

\author{
 Andreas Hadjipieris\footnotemark[1]\enspace\footnotemark[2] \\
  Cyprus Marine and Maritime Institute \\
  Vasileos Pavlou Square 13 \\
  Larnaca, 6023, Cyprus \\
  \And
 Neofytos Dimitriou\footnotemark[1] \\
  Cyprus Marine and Maritime Institute \\
  Vasileos Pavlou Square 13 \\
  Larnaca, 6023, Cyprus \\
  \And
 Ognjen Arandjelović \\
  University of St. Andrews \\
  St Andrews, KY16 9AJ \\
  United Kingdom \\
}
\begin{document}
\maketitle
\footnotetext[1]{Equal contribution.}
\footnotetext[2]{Corresponding author. Email: andreashp@gmail.com}
\begin{abstract}
Port berthing sites are regions of high interest for monitoring and optimizing port operations. Data sourced from the Automatic Identification System (AIS) can be superimposed on berths enabling their real-time monitoring and revealing long-term utilization patterns. Ultimately, insights from multiple berths can uncover bottlenecks, and lead to the optimization of the underlying supply chain of the port and beyond. However, publicly available documentation of port berths, even when available, is frequently incomplete -- e.g.\ there may be missing berths or inaccuracies such as incorrect boundary boxes -- necessitating a more robust, data-driven approach to port berth localization. 
% based on probabilistic modeling of vessel behavior and port activity patterns.
In this context, we propose an unsupervised spatial modeling method that leverages AIS data clustering and hyperparameter optimization to identify berthing sites. Trained on one month of freely available AIS data and evaluated across ports of varying sizes, our models significantly outperform competing methods, achieving a mean Bhattacharyya distance of~\textit{0.85} when comparing Gaussian Mixture Models (GMMs) trained on separate data splits, compared to~\textit{13.56} for the best existing method. Qualitative comparison with satellite images and existing berth labels further supports the superiority of our method, revealing more precise berth boundaries and improved spatial resolution across diverse port environments.
\end{abstract}


% \begin{highlights}
% \item Propose an unsupervised spatial modeling approach for port berth localization. 
% \item Propose a data augmentation strategy that leverages vessel dimensions and heading.
% \item Propose a model selection, hyperparameter tuning, and model optimization method that does not require ground truth.
% \item Address incomplete port berth documentation using freely available AIS data.
% \item  Demonstrate the method’s adaptability across ports of varying sizes and complexity.
% \end{highlights}

%% Keywords



% \end{keyword}

% \end{frontmatter}

% \linenumbers



\section{Introduction}
\label{intro}
The Automatic Identification System (AIS) 
plays a pivotal role in the digitization of the shipping industry by providing frequent vessel movement data~\cite{Yang2019}. An AIS transponder is mandatory for all ships with a gross tonnage over 300 that sail in international waters, ships over 500 tons that do not sail internationally, and all passenger ships~\cite{imo}. With over 310 billion AIS messages transmitted every year, the maritime sector has unquestionably entered the big data era~\cite{louart2023detection}. Originally intended to prevent ship collisions~\cite{Yang2019}, ongoing improvements in data quality and coverage have greatly expanded the potential applications of AIS data.
However, despite accessibility to free AIS data with the appropriate infrastructure (e.g. base stations), organizations that collect and store \textit{global} AIS data typically charge for access, creating a significant barrier to entry, and hindering the big data potential on the maritime sector.
Recently, a few providers, most notably AISStream\footnote{\url{https://aisstream.io/}} and AISHub\footnote{\url{https://aishub.net}}, have started offering real-time terrestrial AIS data at no cost.  
In this work, we focus on open-access sources, which can be sustainably collected across both large spatial and temporal scales. To address the challenge of incomplete port berth documentation, we propose an unsupervised spatial modeling framework that leverages freely available AIS data to identify berthing sites. 
% For our experiments, we use terrestrial AIS data from open-access sources to ensure both scalability and reproducibility.

\begin{figure}[ht]
  \includegraphics[width=\linewidth]{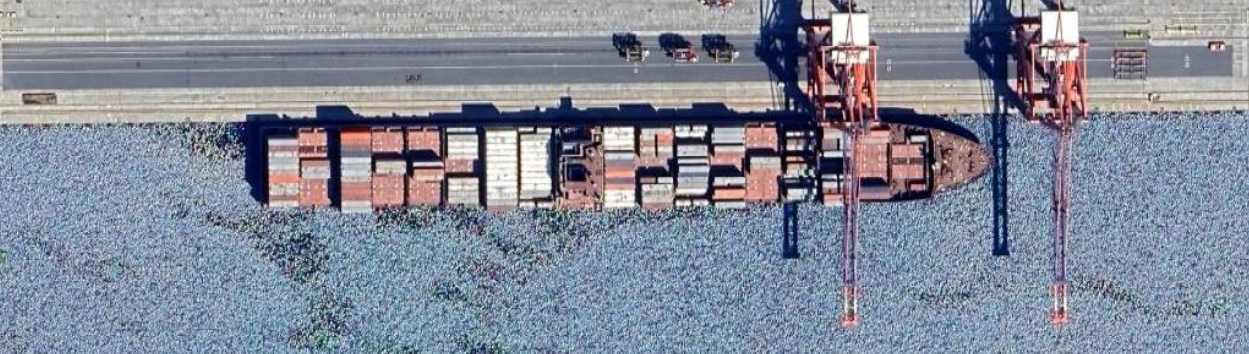}
  \caption{An example of a port berth in the port of Limassol used for loading and unloading container ships. Source: \cite{GoogleMapsImage}}
  \label{fig:boat1}
\end{figure}

Port berths are designated locations within ports equipped with necessary facilities such as cranes and docking infrastructure (see Figure~\ref{fig:boat1}) that support essential maritime operations including cargo loading and unloading, refueling, and maintenance. Despite their critical role, many ports fail to document their location adequately, and existing documentation is often fraught with inaccuracies or omissions. 

By accurately identifying and localizing port berths, we aim to fill these data gaps, ensuring that such vital information is both freely available and reliable. Importantly, port berth localization can enable AIS geofencing, i.e.\ AIS data overlaid on a set of berths to discern short-term and long-term patterns. Examples of applications include the generation of real-time and historical statistics, e.g. service time, the development of predictive data-driven models, and the use of data, statistics, and models by port authorities to make informed decisions regarding infrastructure investments, expansions (e.g.\ new terminals), or operational adjustments~\cite{eskafi2021model}. 
% Such foresight can enable ports to proactively adapt to anticipated industry trends, traffic growth, and evolving regulatory requirements.
% \end{itemize}

The shortage of complete and accurate port berth documentation motivates the importance of port berth localization research. However, existing work on this problem is limited~\cite{SteeLwakNurmTalo+2022,yan2022extracting}.

% \begin{table}[ht]
%   \caption{Selected ports with their respective annual TEUs and the number of AIS messages received from AISStream over a period of 1 month.}%
%   \label{table:ports}
% \centering
% \scalebox{0.9}{

% \begin{tabular}{cccc}
% \hline
% \textbf{Port} & \textbf{Country} & \textbf{Annual TEU (k)} & \textbf{\# of AIS messages} \\ \hline
% Singapore     & Singapore        & 37000                   & 612406                      \\
% Busan         & South Korea      & 22000                   & 195627                      \\
% Antwerp       & Belgium          & 13000                   & 394162                      \\
% Los Angeles   & USA              & 5000                    & 41462                       \\
% Ambarli       & Turkey           & 2600                    & 13                          \\
% Southampton   & UK               & 1600                    & 29290                       \\
% Auckland      & New Zealand      & 818                     & 22814                       \\
% Livorno       & Italy            & 800                     & 56284                       \\
% Cape Town     & South Africa     & 720                     & 40088                       \\
% Gdansk        & Poland           & 685                     & 67492                       \\
% Limassol      & Cyprus           & 348                     & 15631                       \\
% Algeciras     & Spain            & 337                     & 27386                       \\ \hline
% \end{tabular}}
% \end{table}

\setlength{\tabcolsep}{4pt} % Adjust column spacing
\renewcommand{\arraystretch}{0.9} % Adjust row spacing

\begin{table}[ht]
\centering
 \caption{Selected ports with their respective annual TEUs and the number of AIS messages received from AISStream over a period of 1 month.}%
\scalebox{0.84}{
\begin{tabular}{cccc|cccc}
\hline
\textbf{Port}     & \textbf{Country}     & \textbf{TEU (k)} & \textbf{\# of AIS}  & \textbf{Port}      & \textbf{Country}      & \textbf{TEU (k)} & \textbf{\# of AIS}  \\ \hline
Singapore         & Singapore           & 37000            & 612406        & Auckland           & New Zealand          & 818             & 22814        \\
Busan             & South Korea         & 22000            & 195627        & Livorno            & Italy                & 800             & 56284        \\
Antwerp           & Belgium             & 13000            & 394162        & Cape Town          & South Africa         & 720             & 40088        \\
Los Angeles       & USA                 & 5000             & 41462         & Gdansk             & Poland               & 685             & 67492        \\
Ambarli           & Turkey              & 2600             & 13         & Limassol          & Cyprus              & 348              & 15631        \\
Southampton        & UK                   & 1600            & 29290   
% Limassol          & Cyprus              & 348              & 15631        
& Algeciras          & Spain                & 337             & 27386        \\ \hline
\end{tabular}
}

  \label{table:ports}
\end{table}

In this work, we devise an unsupervised method for optimizing data-driven models and demonstrate its efficacy across various ports around the globe. By selecting a diverse set of ports (based on twenty-foot equivalent unit (TEU)), we ensure our sample encapsulates a wide range of port sizes and operational contexts, from small sea ports to some of the world's largest shipping hubs (see Table~\ref{table:ports}). 
The promising results of our method across $11$ ports underscore the proposed method's adaptability and its potential applicability to ports globally. Our contributions are as follows:
\begin{itemize}
    \item We propose a data augmentation strategy based on spatial information that is readily available from AIS data, and experimentally validate its efficacy. This augmentation increases the information regarding the occupying space of a vessel by considering its dimensions and heading.
    \item We introduce a score based on KL-divergence and a hyperparameter optimization evaluation approach and utilize them to explore model selection based on a novel combination of Bayesian optimization and the minimum description length (MDL).
    \item We utilize a post-processing step to translate the underlying distributions of Gaussian Mixture Models (GMMs) to polygons of predicted port berths, and conduct a comparative analysis on port berth localization using as a metric the Bhattacharyya distance based on Monte Carlo Integration (MCI). 
    \end{itemize}

\subsection{Related work}
Related to our work are previous efforts on using AIS data to identify \textbf{stop episodes}~\cite{wang2018enhancing,nogueira2017statistical}. Given that vessels at berths may exhibit a similar behaviour, detecting stop episodes in a port can offer insights on the location of port berths. For example, Wang and McArthur~\cite{wang2018enhancing} and Nogueira et al.~\cite{nogueira2017statistical} developed models that exploit gaps in time and space and detect both stop episodes and differentiating movement patterns in GPS trajectory data. 

In the area of \emph{stay point detection} (i.e.\ where a vessel or vehicle stops) using GPS data, Hwang et al.~\cite{hwang2017detecting} and Luo et al.~\cite{luo2017improved} utilized Density-Based Spatial Clustering of Applications with Noise (DBSCAN), a well-established density-based clustering algorithm, demonstrating its effectiveness in accurately identifying and categorizing stop points within large and noisy datasets.
Further optimization of DBSCAN to handle large spatial datasets efficiently has been achieved through the grid-tree indexing method proposed by Huang et al. \cite{grit}, significantly reducing computational complexity. Huang et al. \cite{grit} introduced GriT-DBSCAN, enhancing DBSCAN's performance by using a grid-tree structure to efficiently index spatial data. Additionally, Chen et al. \cite{prune} demonstrated an optimized variant of DBSCAN tailored for high-dimensional data by pruning unnecessary distance computations, thus significantly speeding up the clustering process.
% in a maritime context
Tang et al.~\cite{TANG2021110108} focused on clustering vessel trajectories by utilizing the Adaptive-threshold Douglas-Peucker and OPTICS algorithms to extract local trajectory features. Their work reaffirms the effectiveness of clustering algorithms in obtaining crucial information, such as stop points, from trajectory data~\cite{TANG2021110108}.
Wang et al. \cite{wang2022support} extended these clustering methodologies specifically for sequential data clustering, employing a support structure representation approach beneficial for identifying patterns in sequential trajectory data. Specifically, Wang et al. \cite{wang2022support} developed a method that captures underlying structural information of sequential data, enabling more accurate clustering of trajectories.
% insights into vessel operation information
% , further enhancing the understanding of maritime activities.

An alternative approach that could be considered for port berth localization is the \emph{spatial clustering} of an area of interest. For example, Millefiori et al.~\cite{millefiori} adapt the Kernel Density Estimation algorithm to the map-reduce paradigm to determine the boundaries of the port of Shanghai. 
% Furthermore, Nicolaou et al.\have conducted research into finding berth congestion and berth occupancy using Poisson's distribution, offering a quantitative approach to managing port resources~\cite{nicolaou1967berth}.
% In the realm of stay point detection (the detection where a vehicle stops) using GPS data, Hwang et al.\ and Luo et al.\have also utilized Density-Based Spatial Clustering of Applications with Noise (DBSCAN) a well-established density-based clustering algorithm, demonstrating its effectiveness in a maritime context~\cite{hwang2017detecting,luo2017improved}. 
In the work by Xiao et al.~\cite{xiao}, DBSCAN and GMMs are utilized to spatially cluster urban areas.  
% We have chosen to replicate the study conducted by Mouzakitis et al.\ and contrast it with our findings.
Yan et al.~\cite{yan2022extracting} proposed a method that integrates trajectory features with geographic scene semantics, employing DBSCAN and Random Forests, for berth identification and ship stopping information extraction from AIS data. Finally, in the work by Steenari et al.~\cite{SteeLwakNurmTalo+2022}, port berths are detected using DBSCAN with vessel statistics subsequently produced. 
Rehman and Belhaouari \cite{divide} further introduced a clustering algorithm that does not require preset cluster counts, dynamically adjusting granularity through iterative splitting and merging. Their algorithm improves clustering adaptability by automatically balancing between dividing and merging clusters based on data-driven criteria.
Our approach also builds upon advances in model-based clustering, notably Gaussian Mixture Models (GMMs), where recent methodologies by Sampaio et al. \cite{reg} proposed techniques for regularization and optimization to automatically determine model complexity and avoid overfitting. Their work introduces novel regularization strategies that improve the robustness and accuracy of clustering results by systematically selecting the optimal number of Gaussian components.

Finally, our evaluation of clustering consistency across different data splits using the Bhattacharyya distance is methodologically supported by foundational similarity metrics such as the centroid index proposed by Rezaei and Fränti \cite{centroididx}, which quantifies cluster similarity, reinforcing robust validation practices in unsupervised clustering contexts. Their centroid index facilitates a reliable comparison between clustering outcomes, crucial for assessing the stability and consistency of clustering solutions.
The method by Steenari et al.~\cite{SteeLwakNurmTalo+2022} was the most relevant and applicable to port berth localization using AIS data and was therefore implemented for comparative analysis (more information is provided in Section~\ref{comp:analysis}.
% Finally, the method proposed by Yan et al.\ is of most relevance to our work have developed a method that significantly advances berth identification and ship-stopping information extraction from AIS data, employing trajectory features integrated with geographic scene semantics with the utilization of DBSCAN and random forest ~\cite{yan2022extracting}. They were able to successfully detect not-reported ports based on the classified port berths and anchorages. 
% Section 2 reviews the relevant literature, highlighting gaps in current port berth identification efforts.

The rest of the paper is organized as follows. In Section \ref{matandmeth}, we detail our method
, including data preprocessing steps, spatial data augmentation strategy, model selection, and the use of GMMs for spatial clustering
. Our experimental findings are detailed in Section \ref{results}.
% , corroborate to the effectiveness of our methodology with significant improvements over existing methods in port berth localization across various port configurations and sizes. 
% Finally, in Section \ref{discussion}, we discuss our findings providing information on where our method excels and furthermore, in which cases our method can be applied other than the maritime domain, and close by interpreting our results into the largest context of port optimization. 
We conclude, in Section \ref{discussion}, where we discuss our findings 
% from the comparative analysis
and contextualize them within the broader scope of port optimization.
% , and highlight the potential and identifying other potential applications beyond the maritime domain. 
% We conclude by contextualizing our results within the broader scope of port optimization.
%%%%%%%%%%%%%%%%%%%%%%%%%%%%%%%%%%%%%%%%%%
\begin{figure}[ht]
    \centering
    \begin{minipage}{0.44\textwidth}
        \centering
        \includegraphics[width=\linewidth, height=\linewidth]{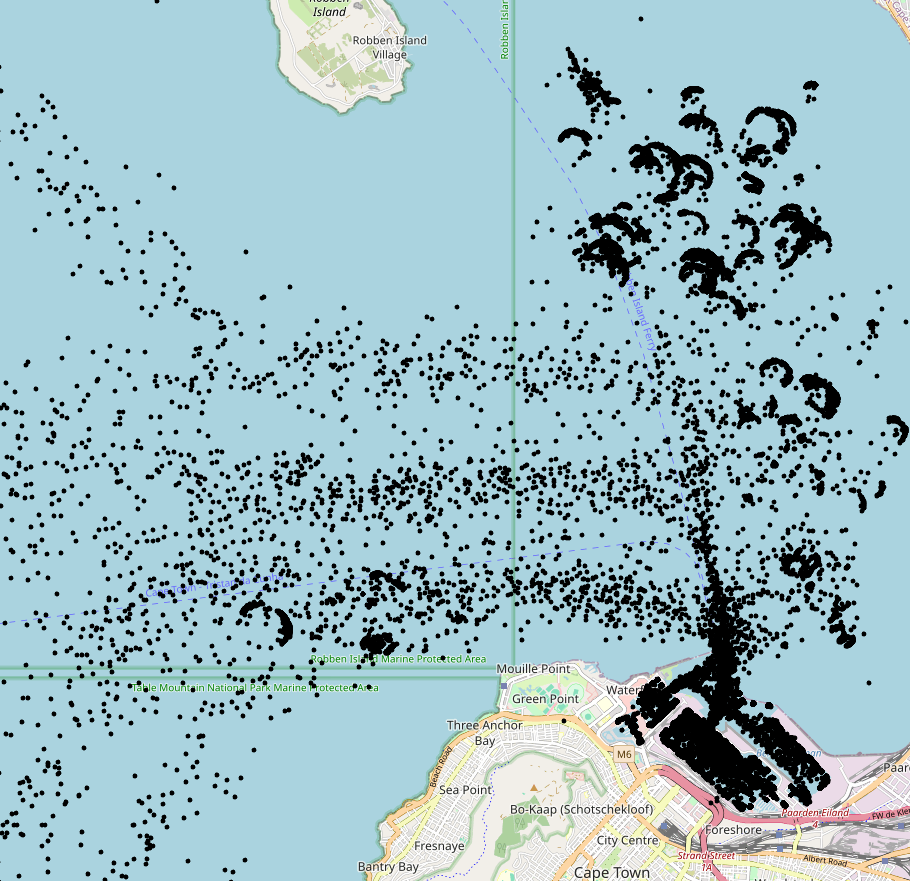}
        \caption{Visualization of the AIS messages (black dots) collected within a month for the port of Cape Town and its surroundings. }
        \label{fig:raw}
    \end{minipage}
    \hfill
    \begin{minipage}{0.44\textwidth}
        \centering
        \includegraphics[width=\linewidth, height=\linewidth]{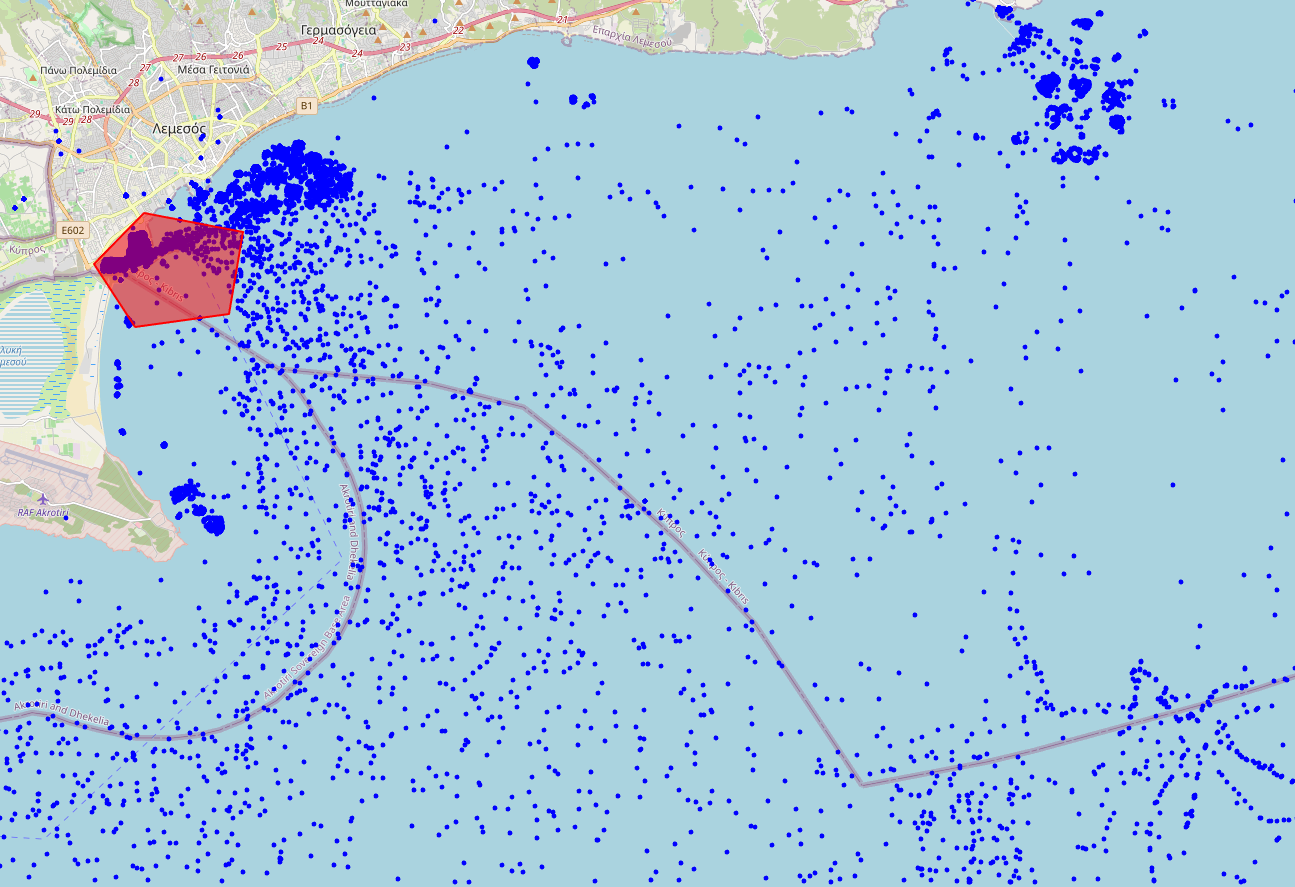}
        \caption{Visualization of the AIS messages (blue dots) collected within a month for the port of Limassol
        (red pentagon).}
        \label{fig:limassol_port_area}
    \end{minipage}
\end{figure}

\section{Materials and Methods}
\label{matandmeth}
\subsection{AIS data}
The AIS data utilized in this research were sourced using the Application Programming Interface (API) of AISStream.io, a platform that offers free access to terrestrial AIS data. Their network of AIS receiving stations, which varies in density globally, covers areas within approximately 200km of the majority of the world's coastlines. For a detailed visualization, a map of their AIS stations is available on their website. To receive AIS data from AISStream.io, one needs to define the region or regions of interest (ROIs) and, optionally, a list of vessels of interest (VOIs) based on Maritime Mobile Service Identities (MMSIs). The MMSI is a nine-digit number used for identifying vessels, ship stations, and coast stations in maritime communication and navigation systems. However, while an MMSI is linked to a single ship at a given time, it can be reassigned to different ships over time, and a ship might change its MMSI. Therefore, the MMSI is not a unique identifier, though it is used as an indicator for individual ships in this work. Finally, we do not utilize the vessel-specific filtering of the API. AIS data from the Cape Town and Limassol ports over a period of one month are shown in Figures~\ref{fig:raw} and~\ref{fig:limassol_port_area}, respectively.

\subsubsection{Regions of interest}

% Stemming from the port-centric nature of our problem, herein,
% Given the port-centric interest of this work,
The port-specific ROI is an area covering an entire port while avoiding adjacent waterways or nearby ports. An example of the Limassol ROI compared to all the data collected for that port is shown in Figure \ref{fig:limassol_port_area}. To ensure a representative sample of ports of all sizes (based on TEU) the following sampling technique was employed. First, twelve port sizes in the range of $100$ to $50,000$ TEU were sampled using
a uniform logarithmic distribution. Ports corresponding closely to these predetermined size categories were then selected (see Table~\ref{table:ports}). 

% Recognizing the complexity of port environments, we defined ROIs to encompass the complete operational areas of each selected port while avoiding adjacent waterways or nearby ports. 
% This precise delineation was crucial for maintaining data integrity, allowing for an accurate assessment of port activities without the confounding influence of external traffic or overlapping zones. Our vigilant approach to defining and verifying ROIs ensures the robustness of our findings, enabling confident extrapolation of our results to broader maritime contexts.

% \subsubsection{Vessels of interest}
% We do not utilize vessel-specific filtering at the API level. 
% However, we only consider AIS data coming from cargo and tanker ships, i.e.\ AIS messages with ``ship type" value in the range of $70$-$89$ inclusively.

\subsubsection{Period of interest}
% While many previous works have employed periods of interest (POIs) of span entire years~\cite{yan2022extracting}, when it comes to the analysis of almost a dozen ports, with some of them being the biggest in the world, e.g. port of Singapore, a POI of an entire year becomes prohibitive for a comprehensive analysis. 
AIS data between the 1st and the 31st of October 2023 were available for the ROIs.
% Should more data become available, we would consider extending the testing period. 
% Despite the short POI, our findings are promissing . indicate that data from one month is adequate for this task. It would be beneficial to conduct further tests to determine if different months yield varying outcomes. 
For the sensitivity analysis on the amount of data (discussed in Section~\ref{results}), the POI varied. Specifically, we carried out four experiments with POIs that begin on the 1st of October and span three days, one week, two weeks, and the entire month, respectively. 
% Longer POIs improve the overlap between port berths and the labels from ShipNext.

% \subsection{Data collection}
% To curate AIS data over the POI, we stored receiving AIS data from the ROIs into a database.
% We collected AIS messages over the ROIs via the API of AISStream.io continually for the above period of time. The number of AIS messages received for each port is given in Table~\ref{table:ports} alongside each port's TEU throughput. Almost no AIS messages were received for the port of Ambarli and were therefore excluded.

The number of AIS messages received over a period of $1$ month for each port is given in Table~\ref{table:ports} alongside each port's TEU throughput. Almost no AIS messages were received for the port of Ambarli which was therefore excluded.

\begin{figure}[ht]
    \centering
    \begin{subfigure}{0.33\textwidth}
  % \centering
        \includegraphics[width=\linewidth]{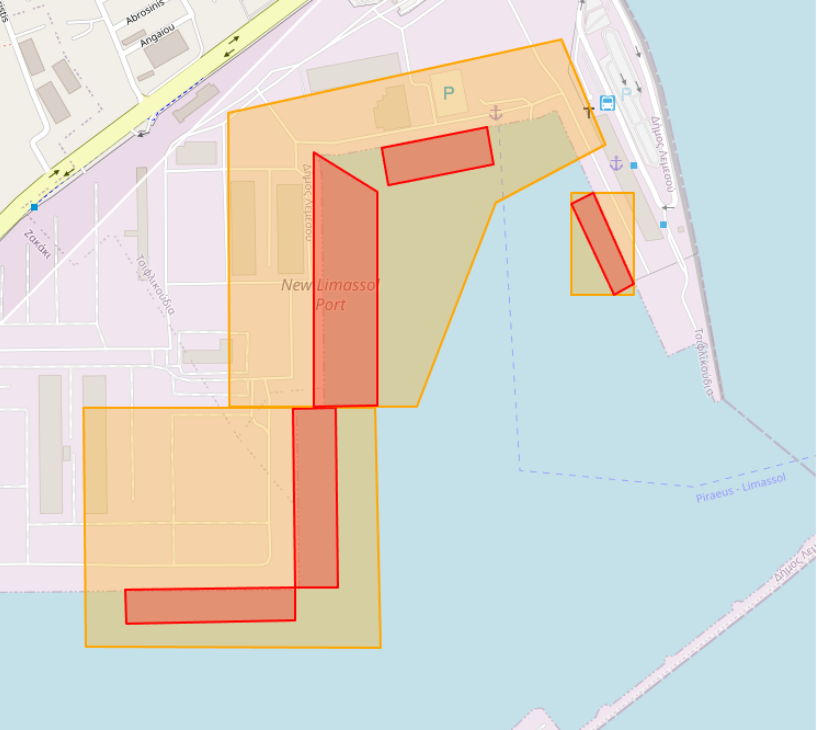}
        \caption{Port of Limassol}
        \label{fig:sub1}
    \end{subfigure}\hfill
    \begin{subfigure}{0.615\textwidth}
  % \centering
        \includegraphics[width=\linewidth]{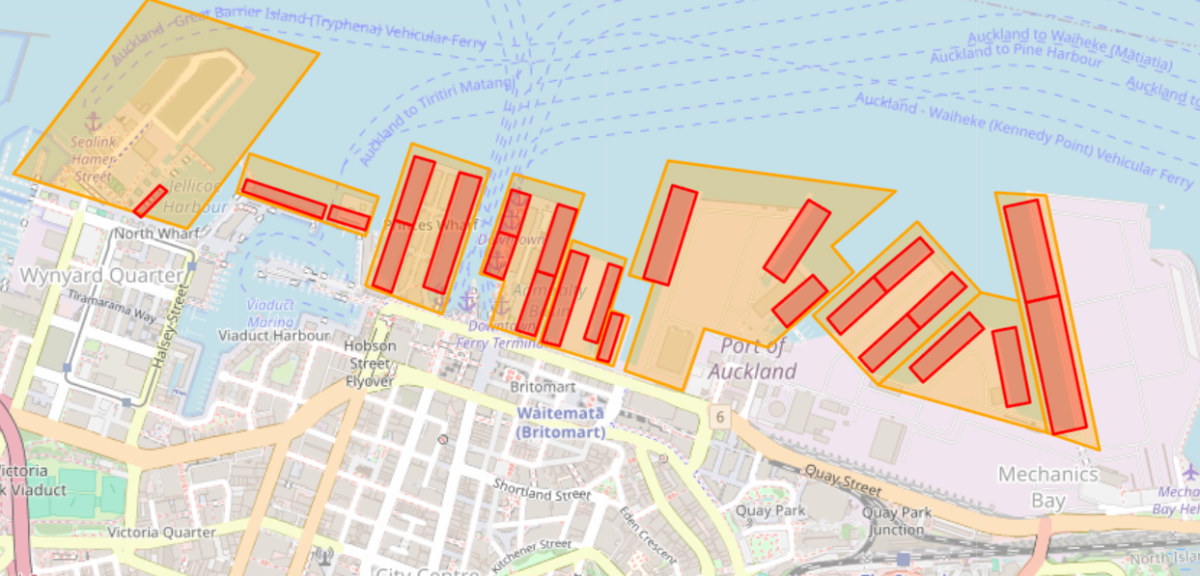}
        \caption{Port of Auckland}
        \label{fig:sub2}
    \end{subfigure}
  \caption{An example of ShipNext labels for the port of Limassol (left) and the port of Auckland (right). The red rotated rectangles are the port berth labels, and the orange areas are the terminal labels.}
  \label{fig:shipnextlabels}
\end{figure}
\subsubsection{Boundary Boxes for Port Berths}
For qualitative model evaluation, rather than training, we collected port berth boundary boxes for the ports of interest. In particular, berth, terminal, and port area labels were obtained from the ShipNext website\footnote{\url{https://shipnext.com/}}. The port berth areas are represented as rotated rectangles as illustrated in Figure~\ref{fig:shipnextlabels}. 
% Notably, as shown e.g.\ in Figure~\ref{fig:vesselai_qualc}, there are inaccuracies in port berth documentation, with many berths not being documented at all.
\subsection{Methodology}
% To systematically address the challenge of port berth localization using AIS data, 
% Our methodology consists of a series of sequential steps, namely data preprocessing, DBSCAN-based filtering, spatial data augmentation, conditional geohash encoding, hyperparameter optimization, Gaussian Mixture Model (GMM) fitting, and evaluation and inference.

% \begin{figure}
%   \includegraphics[trim={0 10cm 12cm 6cm},clip, width=\linewidth]{Figure_5.png}
%   \caption{The data preprocessing pipeline that includes filtering, interpolation, and standardisation.}
%   \label{fig:preprdiag}
% \end{figure}
\subsubsection{Data Preprocessing}
The preprocessing of the raw AIS data involved data cleaning, interpolation, and standardisation. For data cleaning, the data are filtered based on vessel type, so that only cargo and tanker ships remain, and speed ($< 3$ knots). Second, AIS records with either incomplete vessel dimensions or missing heading values are removed. Finally, for each vessel, consecutive messages indicating a substantial directional change in the heading are removed ($> 10$ degrees) since large directional changes may be due to vessel movement or due to noisy or incorrect recorded values. 

Following the above, the AIS messages per MMSI are resampled such that there is at maximum one AIS message per hour (``interpolation'' step). This is done by taking the last message of every hour, if available, for each vessel. An ablation study is carried out to determine the interpolation period as discussed in Section~\ref{section:ablations}. Finally, the latitude and longitude information of all AIS messages are standardized to have a mean of $0$ and a standard deviation of $1$. 
% The preprocessing pipeline is shown in Figure \ref{fig:preprdiag}.

\begin{figure}[ht]
  \includegraphics[width=\linewidth]{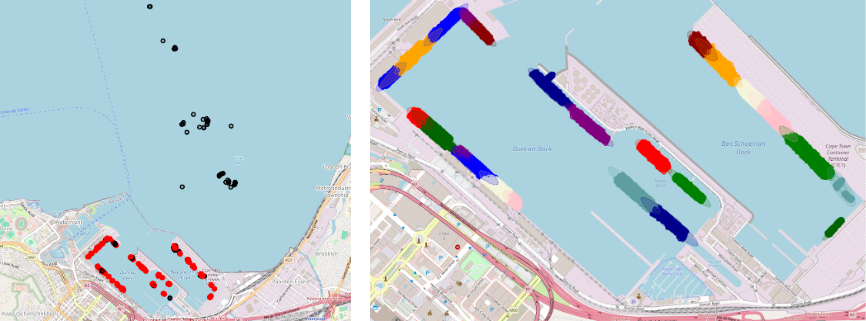}
  \caption{Illustration of DBSCAN algorithm and GMM applied to AIS Data in the port of Cape Town. Left --- The red points indicate AIS data points that are classified as part of clusters, having passed the DBSCAN filtering, while the black points represent those identified as outliers by the algorithm. Right --- The trained GMM with optimal hyperparameters where each Gaussian component represents a potential berth (each Gaussian component is truncated at 3 standard deviations for visualization purposes).}
  \label{fig:dbscan}
\end{figure}

\subsubsection{Data Split} 
For the purposes of hyperparameter tuning and evaluation (but not inference, see Section~\ref{sec:loc}), the preprocessed AIS data for each port is split into two datasets. The splitting process involves the following steps:
\begin{itemize}
    \item Vessels are sorted in descending order based on the number of AIS messages they transmitted.
    \item The sorted vessels are separated into two data sets by alternately assigning consecutive vessels to a different data set.
\end{itemize}

This methodology ensures that the resulting splits of the data have approximately equal numbers of AIS messages while maintaining vessel exclusivity within each split.

\subsubsection{DBSCAN} 
 
Density-Based Spatial Clustering of Applications with Noise (DBSCAN), first introduced by Ester et al.~\cite{dbscan}, is a non-parametric, density-based clustering algorithm, which groups points in a data set based on their proximity and the density of their surrounding points.
DBSCAN is governed by two hyperparameters: \textit{epsilon} $\epsilon$ which defines the maximum radius of a neighborhood, and \textit{minimum points} $p_n$ which refers to the minimum number of points for a cluster to be formed.

In the present work, DBSCAN is applied to each vessel independently, clustering AIS data points based on vessel movement (or lack thereof). Since the AIS data of each vessel are reported hourly, intuitively, the algorithm creates a cluster in locations where the vessel reportedly stayed for at least $p_n$ hours as reported by AIS messages with each reported location being within a distance of $\epsilon$ from one another. Note that haversine distance was used to account for differences in distances between points at different latitudes.
DBSCAN effectively identifies and filters out outliers --- in our case, they can be AIS messages with vessel movement or insufficient duration of stay. Such points are shown as black dots in Figure \ref{fig:dbscan}a. We optimize these two hyperparameters using the Tree Parzen Estimator (TPE) with a log uniform distribution of real numbers between $5$ and $70$ as the prior for $\epsilon$, and a uniform distribution of integers between $2$ and $25$ as the prior for $p_n$.

\begin{figure}[ht]
  \includegraphics[width=\linewidth]{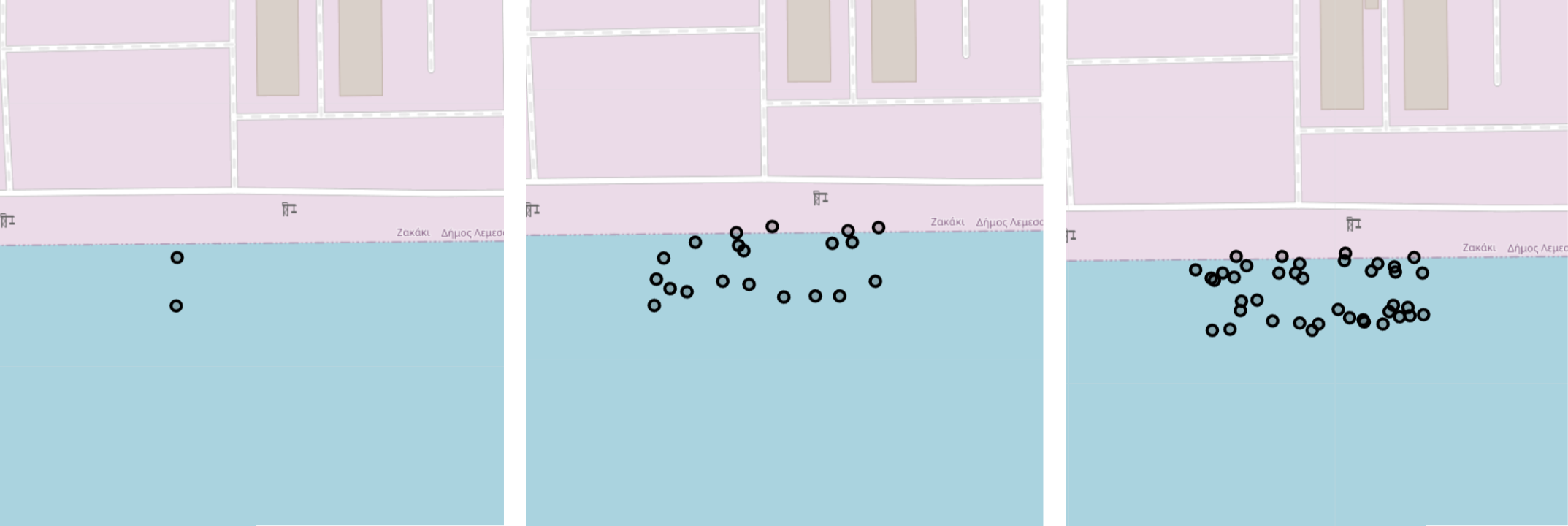}
  \caption{Proposed spatial data augmentation mechanism. Starting with two AIS messages (left), 10 points (mid) and 20 points (right) are generated per AIS message.}
  \label{fig:augmentation}
\end{figure}

% \begin{figure}
% \includegraphics[trim=3cm 0 0 0, width=\linewidth]{Figure_8.png}
%   \caption{Pipeline for augmenting AIS data using DBSCAN, spatial augmentation, and conditional geohash encoding.}
%   \label{fig:augmDBSCANdiagram}
% \end{figure}

\subsubsection{Spatial data augmentation}

Following DBSCAN-based filtering, for each remaining AIS message, we generate additional points ($10$ during training, $20$ during evaluation) by sampling from a uniform distribution within the area occupied by the vessel. This area is determined based on the vessel’s dimensions and heading, which are provided in the AIS data. 
% This information is available in the static messages transmitted by the vessel, which include characteristics that remain constant over time, such as vessel dimensions, call signs, etc.
The dimensions of the vessel are captured in four fields: ``Dimension A'', ``Dimension B'', ``Dimension C'', and ``Dimension D''. These fields report the distances of the AIS receiver to each of the ship's four sides. Based on these dimensions and the reported heading of the vessel, further points are generated as shown in Figure \ref{fig:augmentation}.

\subsubsection{Geohash Encoding}
Following the above, the spatial data of the AIS messages (i.e.~longitude, latitude) can be encoded as geohashes --- a practice that aims to declutter the dataset while significantly improving the efficiency of model training and inference. 
% accelerates the training of the GMM. 
The precision of geohashes is set to $9$ which bins the data into $4.7$m~$\times$~$4.7$m squares. We report on the effectiveness of our proposed method with and without geohash encoding in Section~\ref{results}. 
% The process of using DBSCAN followed by spatial data augmentation and, conditionally, geohash encoding is visualized in Figure~\ref{fig:augmDBSCANdiagram}. 

% \begin{figure}
%   \includegraphics[width=0.9\linewidth]{Figure_9.png}
%   \caption{Finding the optimal ``ncomponents'' value for GMMs based on the smallest (averaged) MDL across a set of candidate values.}
%   \label{fig:gmmdiag}
% \end{figure}

\subsubsection{Gaussian mixture model}
\label{sec:gmm}
A Gaussian Mixture Model (GMM) is a probabilistic model that represents the data as a mixture of several Gaussian distributions, each of which can be thought of as a cluster. GMM does not assign data points to single clusters though; instead, it assigns a probability distribution across all Gaussian components for each data point. 
% This approach enables soft clustering, capturing uncertainty in cluster assignments. For example, GMMs have been used to assist pathfinder algorithms~\cite{data_clustering_gmm}, or in the context of AIS data, GMMs have been used for anomaly detection~\cite{gmm_anomally_ais} and trajectory analysis~\cite{gmm_ais_traj}.
One important hyperparameter in GMMs is the number of components (``ncomponents''), which dictates the number of 
% distinct 
Gaussian distributions underlying the model. In our context, the Gaussian distributions will be interpreted as representing potential berths in a port (illustrated in Figure~\ref{fig:dbscan}b). 
% For visualization purposes, we limit the representation of each Gaussian component to within $3$ standard deviations. 
Therefore, the value of ``ncomponents'' determines how many distinct berths (or other docking areas) a GMM will attempt to model. To determine the optimal ``ncomponents'', we use the Minimum Description Length (MDL) criterion, which can be viewed as a formalization of Occam's razor~\cite{SingAran2022}. It provides a principled approach of balancing model complexity against explanatory power in light of the observed data~\cite{SingAran2022}.
Herein, the description length is the length in bits needed to encode both the parameters of a GMM, and the data 
% (one of the two data splits in our case) 
given the GMM~\cite{SingAran2022}. Formally:
\begin{equation}
DL(M, D) = L(M) + L(D|M)
\end{equation}
where $DL(M, D)$ is the description length corresponding to the model $M$ and data $D$ (given $M$) respectively. The description length can be further elaborated as:
\begin{equation}
DL(M, D) = \frac{1}{2}N_M \log_2 N - \sum_{i=0}^{N-1} \log_2 P(d_i)
\end{equation}
where $d_i$ are individual data points from $D$, $N$ their count, and $N_M$ the number of GMM parameters~\cite{SingAran2022}. The number of parameter $N_M$ of a GMM with $N_C$ ``ncomponents'' can be written as:
% The MDL for each fitted GMM $H$ on data $D$ is calculated as follows:
% \begin{equation}
% \label{eq:mdl}
% MDL(H, D) = \frac{-L(D \mid H) + \frac{1}{2} \cdot L(H) \cdot \log(N_{D})}{N_{D}}
% \end{equation}
% where $L(H \mid D)$ is the log-likelihood of the model $H$ for the data $D$ (or, equivalently, length of encoding data $D$ given model $H$), $N_{D}$ is the number of samples in the data, and $L(H)$ is the number of parameters in the model, calculated as:
\begin{equation}
\label{eq:nparams}
N_M = N_C \times \left(2 \times N_F + \frac{N_F^2 - N_F}{2} + 1\right) - 1
\end{equation}
Here, $N_C$ is the number of mixture components, i.e.\ number of Gaussians in the GMM, and $N_{F}$ is the number of dimensions of the (input) data (in our case it is $2$). Intuitively, Equation~\ref{eq:nparams} accounts for the mean, covariance matrix, and weight parameters that come with every Gaussian.
% Therefore, $N_M$ is equivalent to $5*N_C$.
% INCORRECT! see \url{https://stats.stackexchange.com/questions/229293/the-number-of-parameters-in-gaussian-mixture-model}
This MDL-based approach to tuning ``ncomponents'' balances model complexity with goodness of fit, helping to avoid underfitting (too few berths modeled) or overfitting (e.g.\ modeling noise as berths). 
The process is as follows:
\begin{itemize}
    \item We define a range of possible ``ncomponents'' values: (3, 50) for smaller ports and (30, 250) for larger ports (Singapore and Antwerp in our case).
    \item We fit a GMM with full covariance matrix, $2$ restarts, and $1e^{-4}$ tolerance for each ``ncomponents'' value and for each data split. 
    \item We calculate the MDL for each fitted GMM on each data split. The ``ncomponents'' value yielding the lowest average MDL across the two data splits is selected.
\end{itemize}

% In Figure \ref{fig:gmmdiag} the tuning for the optimal ``ncomponents'' value of the GMM is illustrated.

\subsubsection{Hyperparameter Tuning based on Kullback-Leibler divergence}

The following summarizes our hyperparameter optimization process:
\begin{itemize}
    \item We use the Tree-structured Parzen Estimator (TPE) to search the space of DBSCAN's hyperparameters ($\epsilon$ and $p_n$) for $100$ trials, with the first $30$ serving as a warm start.
    \item At every trial, we fit two GMMs, one for each augmented, as previously defined, data split, using the optimal ``ncomponents'' value (determined as discussed in Section~\ref{sec:gmm}).
    \item We assess the similarity between the two GMMs using the symmetric Kullback-Leibler divergence (KL-div), as described in the equations which follow. 
    \item Following the results on $100$ trials, the trial yielding the lowest KL-div score is selected as the best configuration (i.e.\ optimal $\epsilon$ and $p_n$ for DBSCAN and optimal ``ncomponents'' based on the MDL criterion) for that port. 
\end{itemize}
% We fit two GMMs to different data splits using the optimal ncomponents.
% We assess the similarity between these GMMs using the symmetric Kullback-Leibler divergence (KL-div).
% We use Tree-structured Parzen Estimator (TPE) to efficiently search the hyperparameter space.
% With the optimal number of components for our GMM established, we proceed to fine-tune the remaining hyperparameters. This step is crucial for ensuring the model's robustness across different data splits and its accurate representation of port berth structures.
We use Optuna for hyperparameter optimization~\cite{optuna_2019}. The KL divergence from distribution $p$ to distribution $q$ is defined as:
\begin{equation}
KL(p || q) = \int p(x) \log \frac{p(x)}{q(x)} dx
\label{eq:kl_base}
\end{equation}
To ensure symmetry in our comparison, we use a symmetric variant of KL divergence which we define as:
\begin{equation}
KL_{symm}(p,q) = \max(KL(p || q), KL(q || p))
\label{eq:kl_symm}
\end{equation}
where $p$ and $q$ in our case represent the underlying distribution of GMMs fitted on different data splits.

The symmetric KL divergence quantifies the similarity between GMMs trained on different data splits, reflecting how well the chosen hyperparameters (of both DBSCAN and GMM) generalize across these complementary subsets of the data. Intuitively, a lower KL divergence indicates that the GMMs trained on different data splits identify a similar set of berths, suggesting that the model's representation of port structure is consistent and robust across subsets of the data. The optimal hyperparameter configurations for all ports, automatically selected based on the described process, are listed in~\ref{Optimal hyperparameter values appendix}.
% Pseudocode of the hyperparameter optimization process is provided in Appendix B.

% \begin{figure}
%   \includegraphics[width=\linewidth]{Figure_10.png}
%   \caption{Evaluation approach utilizing the Bhattacharyya distance. }
%   \label{fig:evaldiag}
% \end{figure}
\subsection{Evaluation approach}
\label{sec:eval}

During evaluation, to assess the consistency and robustness of our models across different data splits, we employ the Bhattacharyya Distance (BD) rather than the KL-divergence. For hyperparameter tuning, KL-divergence was more suitable as it more heavily penalizes inconsistencies. In contrast, for evaluation, the Bhattacharyya Distance provides a more balanced assessment (especially since true ground truth is lacking), measuring distribution similarity without overly penalizing minor variations between GMMs trained on different data splits. 
% During evaluation, to assess the consistency and robustness of our models across different data splits, we employ the Bhattacharyya Distance (BD), rather than the KL-divergence. 
% as it qualifies as an absolute metric and can therefore be used to compare different methods (with different data). 
% This approach allows us to quantify the similarity between the probabilistic representations of port structures generated by our models on different subsets of data. 
The evaluation process includes the following steps:
\begin{itemize}
    \item Train GMMs on each of the two augmented data splits (generating $20$ points rather than $10$) using tuned hyperparameters for both DBSCAN and GMM. GMM hyperparameters ``tolerance'' and number of restarts are set to $1e^{-5}$ and $5$ respectively. 
    \item Uniformly sample points from the port area.
    \item Calculate the probability of these points under each of the two GMMs.
    \item Use these probabilities to compute the Bhattacharyya coefficient.
\end{itemize}

The Bhattacharyya coefficient is defined as:
\begin{equation}
BC(P,Q) = \int_{x} \sqrt{p(x)q(x)}dx
\label{eq:bhc}
\end{equation}
where $p(x)$ and $q(x)$ are the values of the probability density functions (PDFs) evaluated at the point x for the two GMMs respectively.
To estimate this coefficient over the entire port area, we use Monte Carlo Integration with a uniform distribution:
\begin{equation}
BC_{MCI}(P,Q) = A\int_{x} \sqrt{p(x)q(x)}\frac{1}{A}dx
\label{eq:MCI}
\end{equation}
where $A$ is the total area of the port.
Finally, we calculate the Bhattacharyya distance by taking the negative logarithm of the coefficient:
\begin{equation}
BD_{MCI}(BC_{MCI}) = -log(BC_{MCI})
\label{eq:nlbd}
\end{equation}
Intuitively, a lower Bhattacharyya distance indicates that the two GMMs assign similar probabilities to the same areas within the port, suggesting that our model consistently identifies similar berth structures across different data splits. For the Steenari et al.\ method~\cite{SteeLwakNurmTalo+2022}, which produces non-probabilistic clusters, we assign a probability of 1 if a sample point falls within a cluster and 0 otherwise, allowing for comparison with our probabilistic approach. To obtain a robust estimate of performance, we calculate the averaged Bhattacharyya distance over $200$ Monte Carlo reruns.

% \begin{figure}
%   \includegraphics[width=\linewidth]{inferencediag.drawio.png}
%   \caption{Inference procedure. }
%   \label{fig:inferdiag}
% \end{figure}

\subsection{Port Berth Localization}
\label{sec:loc}
To localize port berths, the entire data set is used alongside the optimal hyperparameter configuration of both the GMM and DBScan. Moreover, during augmentation, we generate $20$ points rather than $10$. Finally, the hyperparameters of the the GMM, ``tolerance'' and number of restarts, are set to $1e^{-5}$ and $5$ respectively. 
% hyperparameter of GMMs is set to e proposed methodology with tuned hyperpameters nce the optimal hyperparameter configuration is determined for both DBSCAN and GMM, the pipeline is rerun with the following changes:
% \begin{itemize}
%     \item There is no data splitting but rather the pipeline is run on the entire data set.
%     \item The number of generated points per record is increased from $10$ to $20$.
%     \item The tolerance level hyperparameter of the GMMs is set to $0.00001$ and the number of restarts to five, underscoring a shift in priority from efficiency to accuracy.
% \end{itemize}
% Once the fitting process is complete, the model for each port is saved and used in inference mode, as visualized in Figure~\ref{fig:inferdiag}. 
To generate boundary boxes, the augmented AIS points that fall in each component of the final GMM (of each port) are encircled by a rotated rectangle, representing the area of the localized port berth. The rotated rectangle is calculated by finding the smallest possible rectangle that can contain all points in a cluster. To achieve this, we calculate the convex hull of the cluster points and then evaluate the bounding rectangles at different angles formed by the edges of the hull. The rectangle with the smallest area is selected as the final bounding box of each Gaussian component.

% \begin{landscape}
\begin{table}[ht]

\caption{The mean and standard deviation of the Bhattacharyya distance over 200 reruns of our method with Geohash and without Geohash applied, and of the method by Steenari et al.\ applied on AIS data of one month. Notably, the method by Steenari et al.\ on the Port of Los Angeles returned $\infty$ and was therefore excluded.}
\scalebox{0.93}{
\begin{tabular}{cccccccccc}
\hline
\textbf{}            &  & \textbf{Non-Geohash} &              &           & \textbf{Geohash} &              &           & \textbf{Steenari et al.} &              \\ \hline
\textbf{Port}        &  & \textbf{Mean}        & \textbf{STD} & \textbf{} & \textbf{Mean}    & \textbf{STD} & \textbf{} & \textbf{Mean}            & \textbf{STD} \\ \hline
\textbf{Singapore}   &  & 0.882                & 0.057        &           & \textbf{0.850}   & 0.042        &           & 9.979                    & 0.191        \\
\textbf{Busan}       &  & 0.800                & 0.143        &           & \textbf{0.687}   & 0.135        &           & 14.870                   & 0.776        \\
\textbf{Antwerp}     &  & 1.145                & 0.071        &           & \textbf{0.957}   & 0.057        &           & 10.383                   & 0.133        \\
\textbf{Los Angeles} &  & 0.950                & 0.046        &           & \textbf{0.855}   & 0.035        &           & -                        & -            \\
\textbf{Southampton} &  & 0.985                & 0.199        &           & \textbf{0.672}   & 0.132        &           & 16.887                   & 0.680        \\
\textbf{Auckland}    &  & 1.237                & 0.258        &           & \textbf{0.922}   & 0.193        &           & 12.487                   & 0.437        \\
\textbf{Livorno}     &  & 1.020                & 0.340        &           & \textbf{0.830}   & 0.236        &           & 15.458                   & 0.507        \\
\textbf{Cape Town}   &  & 0.815                & 0.082        &           & \textbf{0.618}   & 0.066        &           & 16.751                   & 0.511        \\
\textbf{Gdansk}      &  & 1.235                & 0.084        &           & \textbf{0.955}   & 0.057        &           & 14.736                   & 0.817        \\
\textbf{Limassol}    &  & 0.734                & 0.080        &           & \textbf{0.705}   & 0.060        &           & 13.192                   & 0.501        \\
\textbf{Algeciras}   &  & 1.288                & 0.088        &           & \textbf{1.142}   & 0.053        &           & 13.058                   & 0.518        \\
                     &  &                      &              &           &                  &              &           &                          &              \\ \hline
\textbf{Average}     &  & 1.008                & 0.132        &           & \textbf{0.836}   & 0.097        &           & 13.780                   & 0.507        \\ \hline
\end{tabular}
}
% \end{tabularx}
% \end{adjustwidth}

\label{tab:restable}
\end{table}
% \end{landscape}

%%%%%%%%%%%%%%%%%%%%%%%%%%%%%%%%%%%%%%%%%%
\section{Results}
\label{results}

\subsection{Comparative Analysis}
\label{comp:analysis}
In this section we conduct a comparative analysis between the proposed method and the method by Steenari et al.~\cite{SteeLwakNurmTalo+2022} on port berth localization with one month of AIS data. For our method, the number of AIS messages that remain after preprocessing (but before DBSCAN and augmentation) are: Algeciras ($2886$), Antwerp ($23309$), Auckland ($1268$), Busan ($10380$), Cape Town ($2563$), Gdansk ($8070$), Limassol ($403$), Livorno ($3226$), Los Angeles ($2721$), Singapore ($23521$), and Southampton ($1762$). Further information is provided in ~\ref{Data preprocessing appendix}. For the competing method, based on the preprocessing used by Steenari et al., the number of AIS messages that remain before DBSCAN are: Algeciras ($7572$), Antwerp ($160368$), Auckland ($2784$), Busan ($45030$), Cape Town ($40307$), Gdansk ($19056$), Limassol ($1051$), Livorno ($7173$), Los Angeles ($5625$), Singapore ($59313$), and Southampton ($4146$). Notably, for all ports, our method is superior even though it consistently retains less than half of the amount of AIS messages than the competing method. 
% Note that this number represents the average number of AIS messages of the data splits.
%%%%%%%%

\begin{figure}[ht]
% \begin{adjustwidth}{\linewidth}{0cm}
\centering
\includegraphics[width=\linewidth]{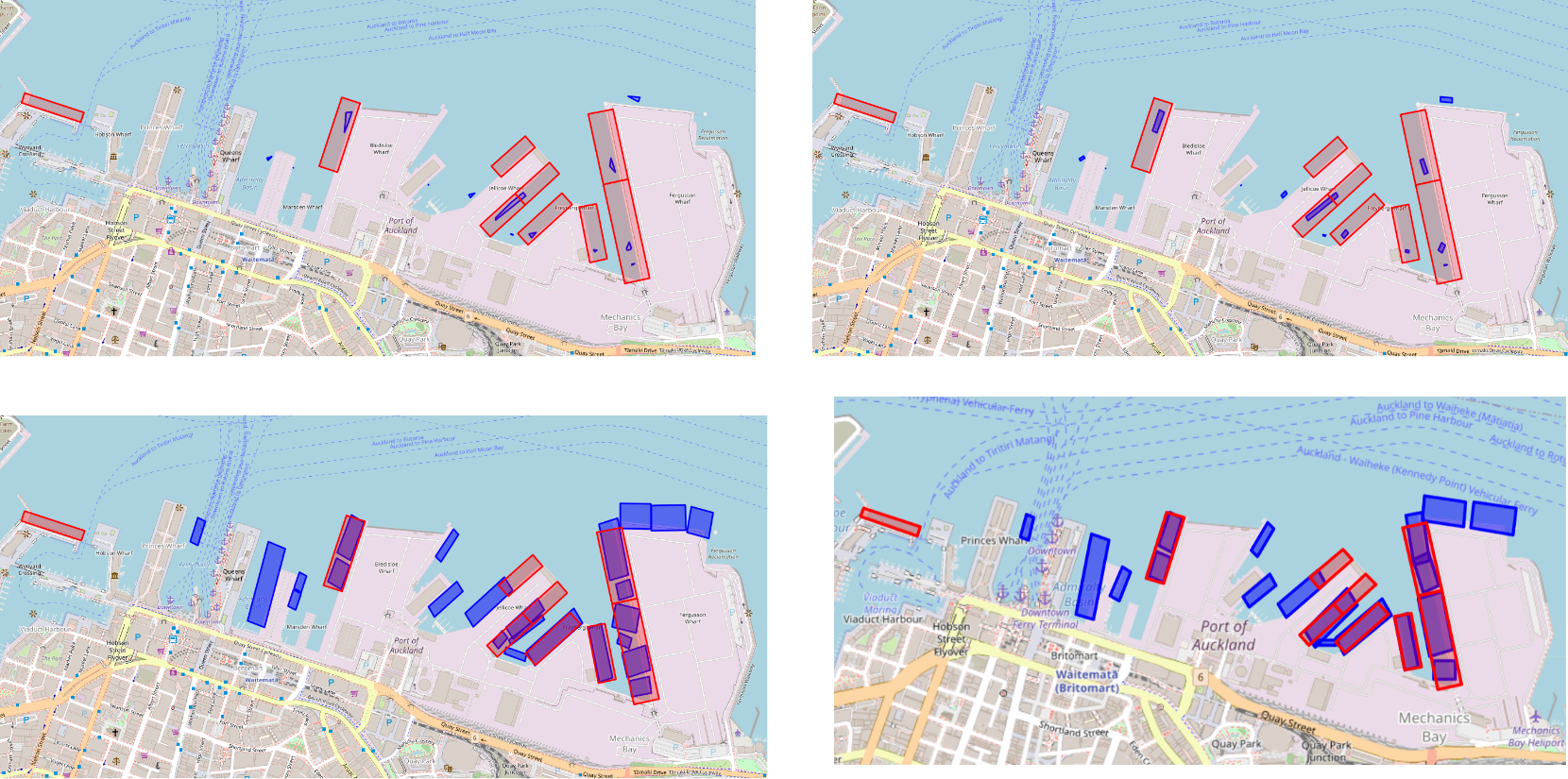}
% \end{adjustwidth}
\caption{Comparison between original competing method (Top Left), competing method augmented with rotated rectangles (Top Right), proposed method without geohash encoding (Bottom Left), and proposed method with geohash encoding (Bottom Right) applied to the port of Auckland. Blue areas represent the predictions while red areas represent the labels.}
 \label{fig:vesselai_qualc}
\end{figure}

The proposed method by Steenari et al.~\cite{SteeLwakNurmTalo+2022} begins by filtering AIS data to records within a specified port polygon (ShipNext's port labels are used in our experiments). Next, data points with a navigational status of $5$ (mooring) and a speed greater than $1$ are removed. Any duplicates based on MMSI and timestamp are also removed. 
% The AIS data is then sorted by MMSI and timestamp to maintain a chronological order. 
Continuous mooring events are subsequently identified where each event is defined as a continuous stream of AIS messages with navigational status set to mooring, lasting more than $> 1$ hour. The AIS messages corresponding to an event are assigned a unique berth number. Next, data points where the navigational status is not set to $5$ are removed. For each event, the median longitude and latitude are calculated to determine the center coordinates of the berths, which are then passed to the DBSCAN algorithm with $\epsilon=50$m and $p_n=3$ for clustering purposes. Finally, convex hull polygons are created for each cluster. For post-processing, we incorporate an additional step by turning the clusters into rotated rectangles. This addition enhances the outcome as can be seen in~Figure~\ref{fig:vesselai_qualc}.

Results of our proposed method, both with and without geohash encoding, as well as the competing method of Steenari et al.\ can be found in Table~\ref{tab:restable}.
% while the results of the periods 3 days, 1 week, and 2 weeks can be found in Appendix . 
The method by Steenari et al.\ scores (\textit{Mean: $13.780$, Std: $0.507$}) consistently worse than the proposed method with (\textit{Mean: $0.836$, Std: $0.097$}) and without geohash encoding (\textit{Mean: $1.008$, Std: $0.132$}). This difference in performance is also visually evident in Figure~\ref{fig:vesselai_qualc}, where the predictions from our proposed method, especially with geohash encoding, resemble more realistic berthing site configurations, producing coherent and practical berth placements. Moving on to a comparison between our two variants, Figure~\ref{fig:vesselai_qualc} highlights that the geohash-enabled variant produces more plausible outcomes than the non-geohash variant. For instance, the geohash-enabled approach avoids the unlikely scenario of identifying 9 berths on the right side of the port which is seen in the non-geohash variant. This improvement in realism demonstrates the effectiveness of incorporating geohash encoding, leading to more accurate and applicable port berth localization. Importantly, these qualitative improvements are consistent with the quantitative results, further reinforcing the superiority of the geohash-enabled approach.

% Interestingly, while there is a clear positive correlation between AIS messages and performance for our method with geohash encoding ($0.27$), that is not the case for the non-geohash variant and the competing method ($0.04$ and $-0.36$ respectively). When we exclude the effects of TEU in the correlation (i.e.\ the residuals from a linear regression of AIS messages on TEU are used instead), the positive correlation remains for the proposed method with geohash ($0.26$) while the non-geohash variance shows almost no correlation ($-0.05$), and the competing method a strong negative correlation ($-0.54$). The standard deviation of the Geohash method compared with the non-Geohash is consistently significantly lower. 
% A qualitative comparison between our method and ShipNext's labels is shown in Figure~\ref{fig:satellite_cape}. As can be seen, we find cases where our model localized berths that were either not part of the ShipNext's labels (b, d, e), or were not accurate enough (f). There were also two cases (a, c) where the opposite was observed - berths part of the ShipNext's labels were not identified by our model. We investigated both (a) and (c) further and have identified the reason to be that that there was no data from cargo and tanker ships during the POI. Further inquiring showed that (a) is primarily used for cruise ships, whereas (c) is used by mining vessel which although fall under the cargo category, the use of the berth is either seasonal or only begun post the POI.

\begin{figure}[ht]
\includegraphics[width=\linewidth]{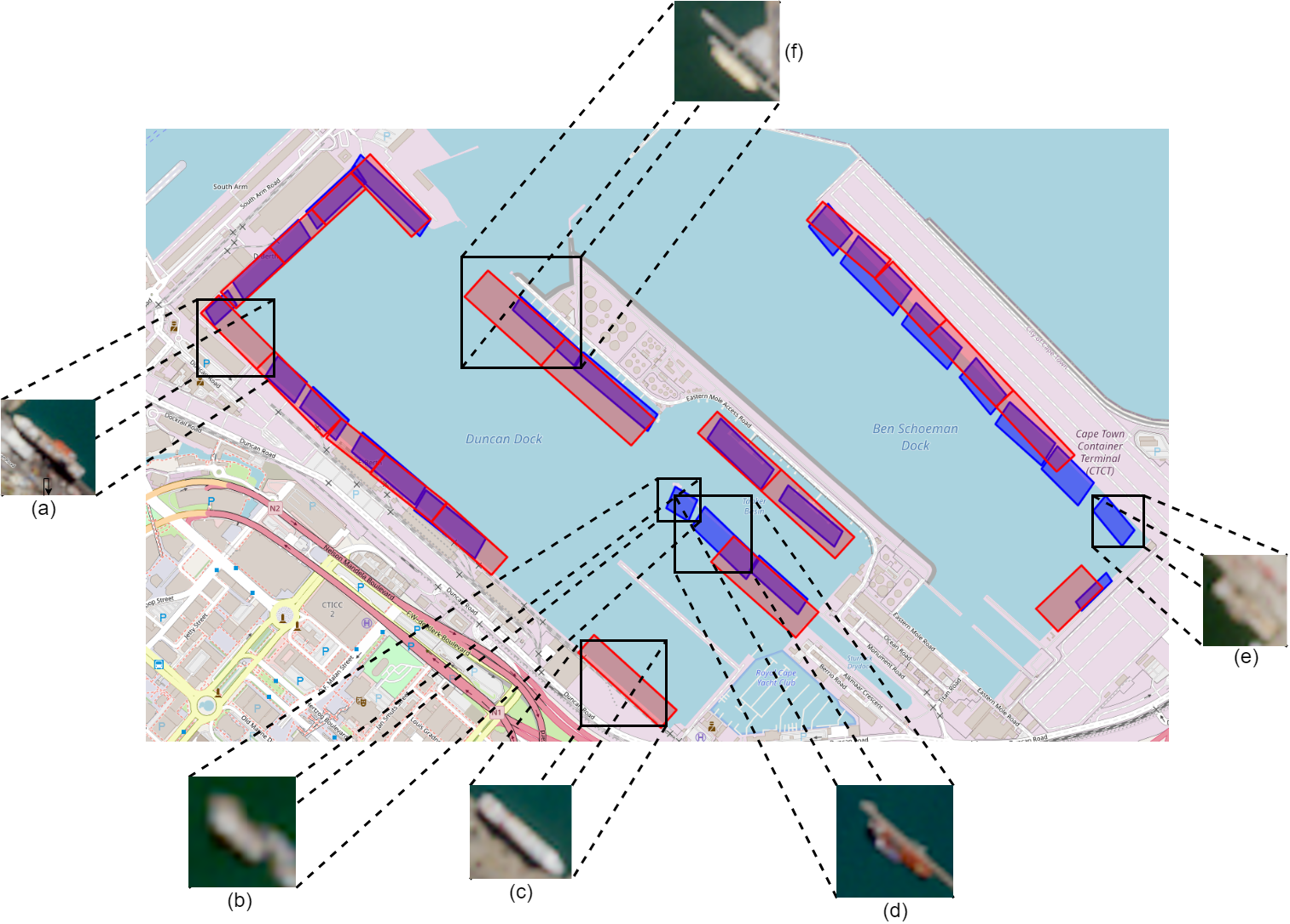}
\caption{Red rectangles are the labels of ShipNext and blue rectangles are the predictions of our model for the port of Cape Town. For each of the black squares, a satellite image is provided. (a, c): Examples where our model did not identify a berthing site whereas ShipNext labels did. (b, d, e):  Examples where our model localized a berthing site but was not documented by the labels from ShipNext. (f): An example where our model provided a more accurate representation of where the berthing site when compared to the ShipNext labels.}
\label{fig:satellite_cape}
\end{figure}

Figure~\ref{fig:satellite_cape} presents a qualitative comparison between our method and the ShipNext's labels. The depicted satellite images were identified from the Sentinel-2 optical archive by starting at the most recent (cloud-free) images and moving backwards in time (up to a year prior or until a ship was found at the region of interest). Our model demonstrates superior performance in several instances, identifying berths that are either omitted from ShipNext's labels (b, d, e) or inadequately delineated (f). However, in two cases (a, c), our model fails to detect berths that are included in ShipNext's labels. Upon investigation of these discrepancies, we find that no cargo or tanker emitted AIS data from the (a) and (c) regions for the POI. Additional investigation reveals that berth (a) is predominantly used by passenger/cruise ships and fishing vessels (ship types outside the scope of our investigation), while berth (c) serves mining vessels. Although mining vessels typically fall under the cargo category, the utilization of berth (c) appears to be either seasonal or to have only commenced after the POI (as confirmed by newer AIS data that indeed document mining vessel activity at this berth).

\subsection{Ablations}
\label{section:ablations}
We report on a number of ablations studies carried out in pursue of examining the importance of different components and characteristics of our data and method. For brevity, this section focuses on findings for the geohash-enabled method, which demonstrated consistently superior performance, while results for the non-geohash version are included in ~\ref{Further ablation results appendix}. 
\subsubsection{Period of interest (POI)}
% \begin{table}[ht]
% \caption{Ablation study results on the effect of the period of interest (3 days, 1 week, 2 weeks, and 1 month) for all ports in consideration. Performance is measured using the Bhattacharyya distance.}
% \label{tab:combined_geohash_results}
% \centering
% \scalebox{0.8}{
% \begin{tabular}{lcccc}
% \hline
% \textbf{Port} & \textbf{3 days} & \textbf{1 week} & \textbf{2 weeks} & \textbf{1 month} \\
% \hline
% Algeciras   & 3.615 & 2.046 & 1.475 & \textbf{1.142} \\
% Antwerp     & 1.838 & 1.381 & 1.169 & \textbf{0.705} \\
% Auckland    & 2.456 & 1.157 & 1.135 & \textbf{0.955} \\
% Busan       & 1.246 & 0.959 & 0.790 & \textbf{0.618} \\
% Cape Town   & 1.617 & 0.968 & 1.162 & \textbf{0.830} \\
% Gdansk      & 3.064 & 1.611 & 1.203 & \textbf{0.922} \\
% Limassol    & 1.137 & 0.940 & 0.763 & \textbf{0.672} \\
% Livorno     & 1.326 & 1.076 & 1.065 & \textbf{0.855} \\
% Los Angeles & 6.791 & 1.521 & 1.060 & \textbf{0.957} \\
% Singapore   & 1.303 & 1.039 & 0.878 & \textbf{0.687} \\
% Southampton & 2.020 & 1.105 & 0.877 & \textbf{0.850} \\
% \hline
% \end{tabular}
% }
% \end{table}

\begin{table}[ht]
\centering
\caption{Ablation study results for the geohash-enabled method showing the effect of the period of interest on performance (Bhattacharyya distance). Left and right columns represent different sets of ports.}
\scalebox{0.8}{
\begin{tabular}{lcccc|lcccc}
\hline
\textbf{Port}     & \textbf{3 days} & \textbf{1 week} & \textbf{2 weeks} & \textbf{1 month} & \textbf{Port}     & \textbf{3 days} & \textbf{1 week} & \textbf{2 weeks} & \textbf{1 month} \\ \hline
                  &                 &                 &                  &                  & Gdansk           & 3.064           & 1.611           & 1.203            & \textbf{0.922}  \\
Algeciras         & 3.615           & 2.046           & 1.475            & \textbf{1.142}  & Limassol         & 1.137           & 0.940           & 0.763            & \textbf{0.672}  \\
Antwerp           & 1.838           & 1.381           & 1.169            & \textbf{0.705}  & Livorno          & 1.326           & 1.076           & 1.065            & \textbf{0.855}  \\
Auckland          & 2.456           & 1.157           & 1.135            & \textbf{0.955}  & Los Angeles      & 6.791           & 1.521           & 1.060            & \textbf{0.957}  \\
Busan             & 1.246           & 0.959           & 0.790            & \textbf{0.618}  & Singapore        & 1.303           & 1.039           & 0.878            & \textbf{0.687}  \\
Cape Town         & 1.617           & 0.968           & 1.162            & \textbf{0.830}  & Southampton      & 2.020           & 1.105           & 0.877            & \textbf{0.850}  \\
\hline
\end{tabular}
}

\label{tab:combined_geohash_results}
\end{table}

\begin{figure}[ht]
\centering
    % \includegraphics[width=0.24\textwidth,height=85pt]{custom_cape_3days (1).png}
    % \includegraphics[width=0.24\textwidth,height=85pt]{custom_cape_1week (1).png}
    % \includegraphics[width=0.24\textwidth,height=85pt]{custom_cape_2weeks (1).png}
    % \includegraphics[width=0.24\textwidth,height=85pt]{custom_cape_1month (1).png}
    % \\
    \includegraphics[width=0.42\linewidth]{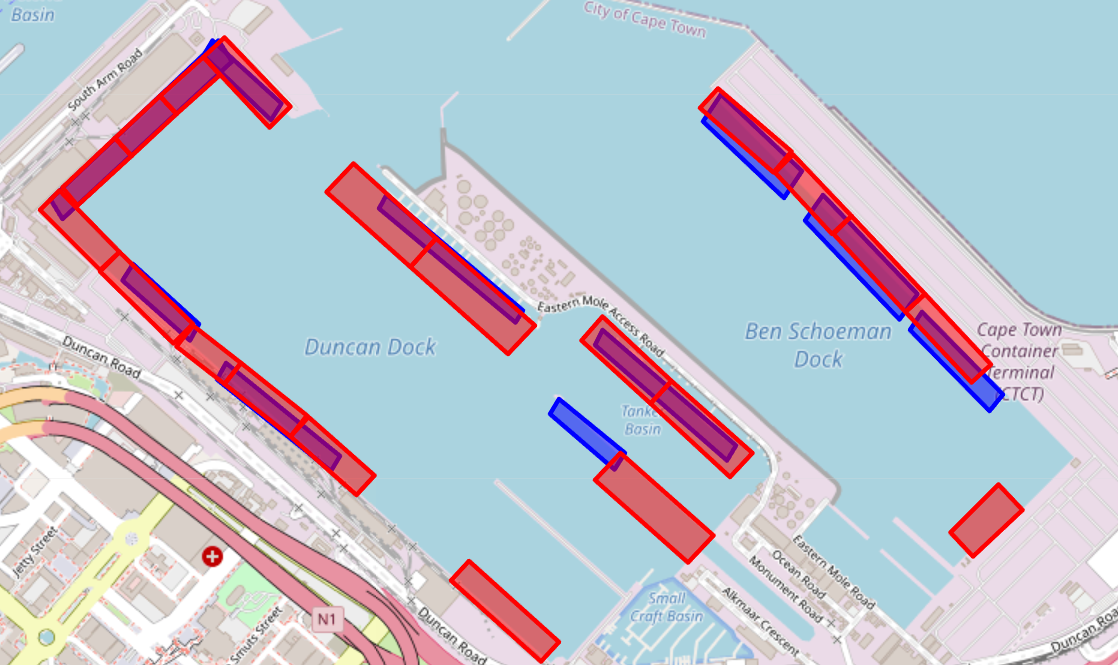}
    \includegraphics[width=0.42\linewidth]{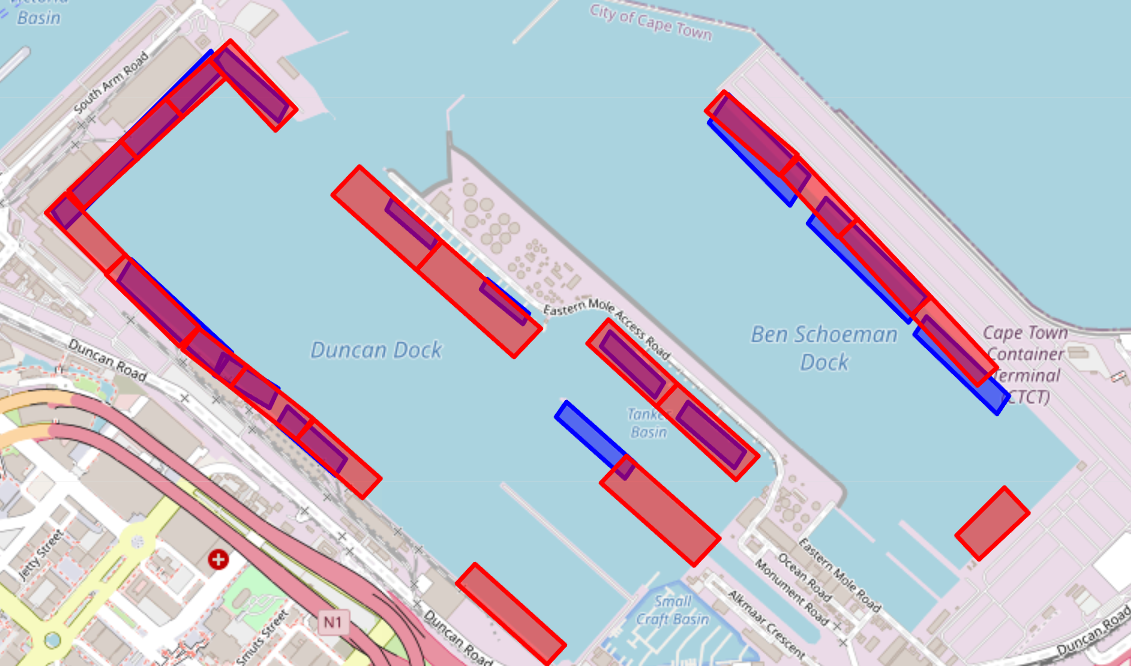}\\
    \includegraphics[width=0.42\linewidth]{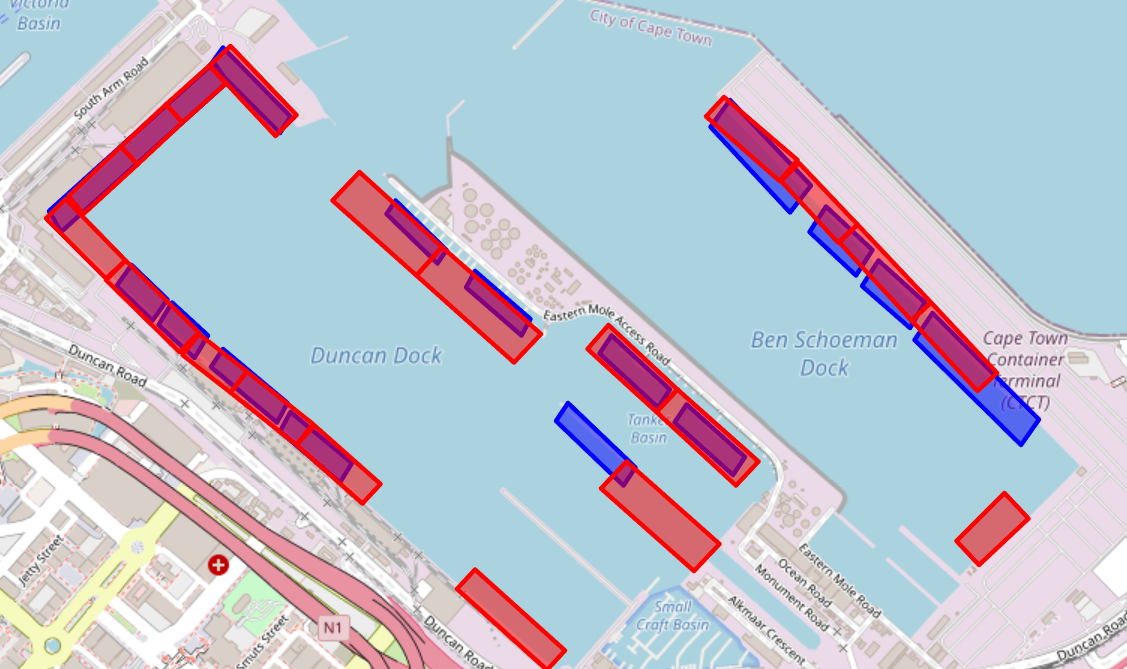}
    \includegraphics[width=0.42\linewidth]{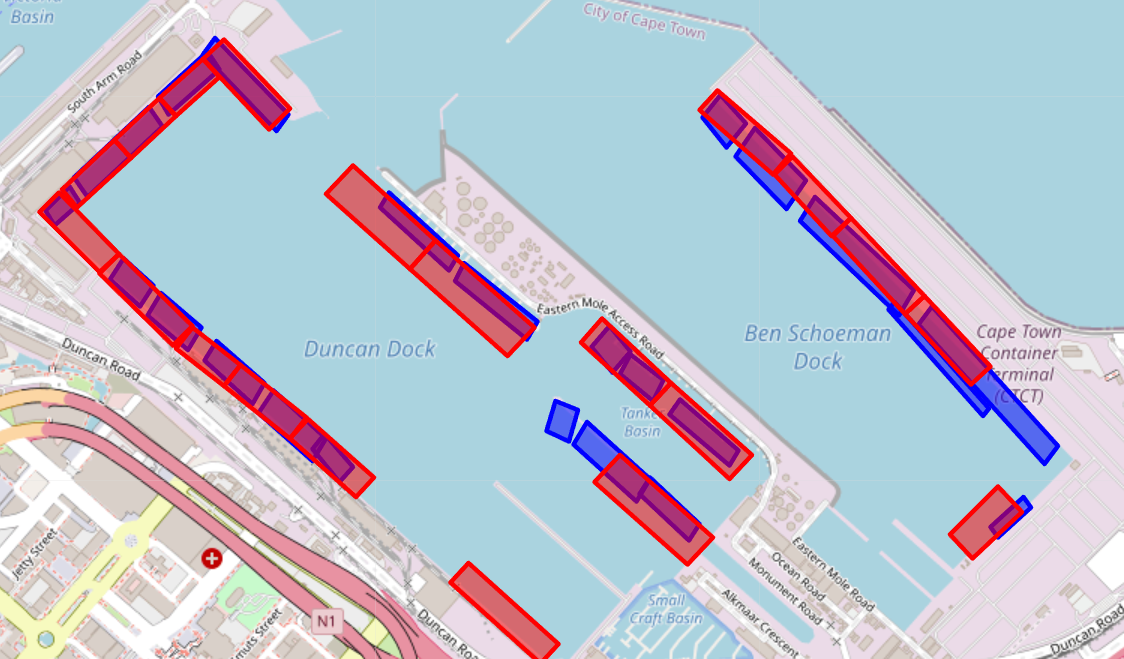}
    
\caption{Qualitative comparison of the proposed geohash-enabled variant with increasing period of interest for the port of Cape Town. From top to bottom and left to right the period is 3 days, 1 week, 2 weeks, and 1 month.}
\label{fig:cape}
\end{figure} 

As shown in Table~\ref{tab:combined_geohash_results}, extending the POI duration improves the consistency of GMM outputs across the two data splits. A qualitative comparison further supporting this observation is presented in Figure~\ref{fig:cape}. This outcome aligns with intuition: longer observation windows yield more AIS messages, which allows for stricter DBSCAN hyperparameters without excessively pruning the dataset. As a result, both false positives (non-berth areas identified as berths) and false negatives (actual berths not detected) are less likely, and berth boundaries are more clearly delineated. 
% Similar results for the non-geohash variant are provided in Appendix B. 
These findings underscore the importance of using sufficient data to ensure robustness in berth localization.

\begin{table}[ht]
\centering
\caption{Ablation study results for the Port of Cape Town. Left: Impact of spatial augmentation (number of generated points) on geohash-based methods. Right: Sensitivity to different interpolation periods. Performance is measured using the Bhattacharyya distance. L-CI and U-CI stand for lower and upper confidence interval bounds ($95\%$), respectively, and STD stands for standard deviation.}
\scalebox{0.85}{
\begin{tabular}{lcccc|lcccc}
\hline
\textbf{\# Generated Points} & \multicolumn{1}{l}{\textbf{Mean}} & \multicolumn{1}{l}{\textbf{L-CI}} & \multicolumn{1}{l}{\textbf{U-CI}} & \multicolumn{1}{l|}{\textbf{STD}} & \textbf{Interpolation} & \multicolumn{1}{l}{\textbf{Mean}} & \multicolumn{1}{l}{\textbf{L-CI}} & \multicolumn{1}{l}{\textbf{H-CI}} & \multicolumn{1}{l}{\textbf{STD}} \\
% \textbf{Generated Points} & \multicolumn{1}{l}{\textbf{L-CI}} & \multicolumn{1}{l}{\textbf{U-CI}} & \multicolumn{1}{l}{\textbf{L-CI}} & \multicolumn{1}{l|}{\textbf{U-CI}} & \textbf{Period} & \multicolumn{1}{l}{\textbf{L-CI}} & \multicolumn{1}{l}{\textbf{U-CI}} & \multicolumn{1}{l}{\textbf{L-CI}} & \multicolumn{1}{l}{\textbf{U-CI}} \\
\hline
\textbf{0 points} & 1.230 & 1.167 & 1.294 & 0.645 & & & & & \\
\textbf{2 points} & 0.657 & 0.650 & 0.665 & 0.077 & \textbf{15 minutes} & 0.621 & 0.612 & 0.630 & 0.067 \\
\textbf{5 points} & 0.654 & 0.644 & 0.664 & 0.072 & \textbf{30 minutes} & 0.616 & 0.606 & 0.626 & 0.069 \\
\textbf{10 points} & 0.641 & 0.631 & 0.650 & 0.068 & \textbf{1 hour} & 0.619 & 0.610 & 0.629 & 0.069 \\
\textbf{20 points} & 0.625 & 0.615 & 0.634 & 0.067 & \textbf{2 hours} & 0.632 & 0.623 & 0.641 & 0.067 \\
\textbf{40 points} & 0.621 & 0.615 & 0.627 & 0.065 & & & & & \\
\hline
\end{tabular}
}

\label{tab:ablation_gen}
\end{table}

%     \begin{figure}
% % \begin{adjustwidth}{-\extralength}{0cm}
% \centering
%     \includegraphics[width=0.3\textwidth,height=95pt]{custom_1.png }
%     \includegraphics[width=0.3\textwidth,height=95pt]{custom_2.png }
%     \includegraphics[width=0.3\textwidth,height=95pt]{custom_5.png }
%     \\
%     \includegraphics[width=0.3\textwidth,height=95pt]{custom_10.png}
%     \includegraphics[width=0.3\textwidth,height=95pt]{custom_20.png}
%     \includegraphics[width=0.3\textwidth,height=95pt]{custom_40.png}
    
% % \end{adjustwidth}
% \caption{Qualitative results for the ablation study of interpolation periods for the port of Cape Town with the Non-Geohash method. From top to bottom and left to right the generated points are 1, 2, 5, 10, 20, 40.}
% \label{fig:gen_custom_ablation_qual}
% \end{figure}
\subsubsection{Number of points generated during spatial augmentation}
This ablation study examines how varying the number of generated points per AIS message during spatial augmentation affects berth boundary precision and clustering consistency. We evaluated configurations with $0$, $2$, $5$, $10$, $20$, and $40$ additional points per message. Table~\ref{tab:ablation_gen} presents the corresponding Bhattacharyya distances for the port of Cape Town.

\begin{figure}[ht]
% \begin{adjustwidth}{-\extralength}{0cm}
\centering
    \includegraphics[width=0.32\linewidth]{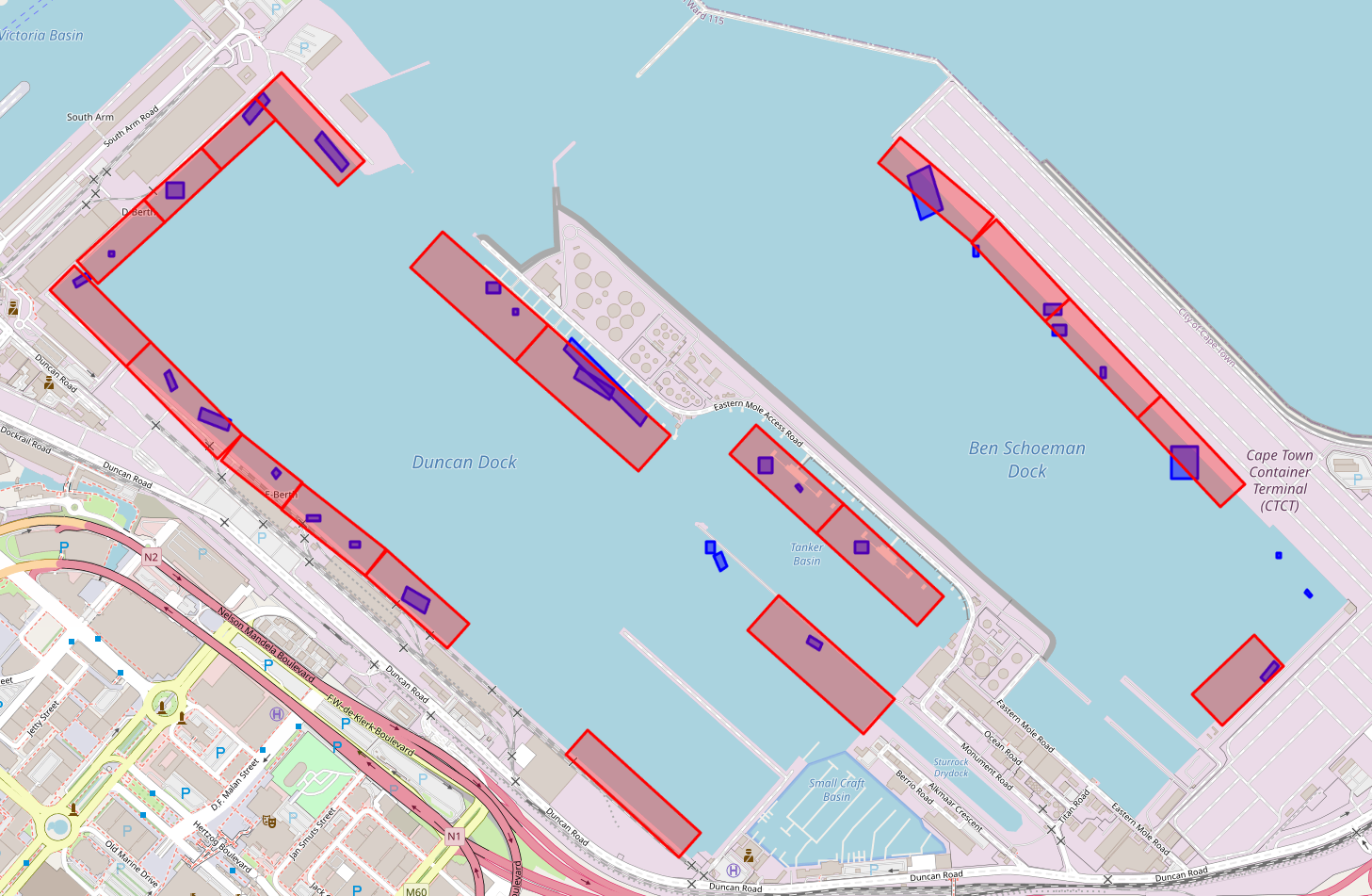 }
    \includegraphics[width=0.32\linewidth]{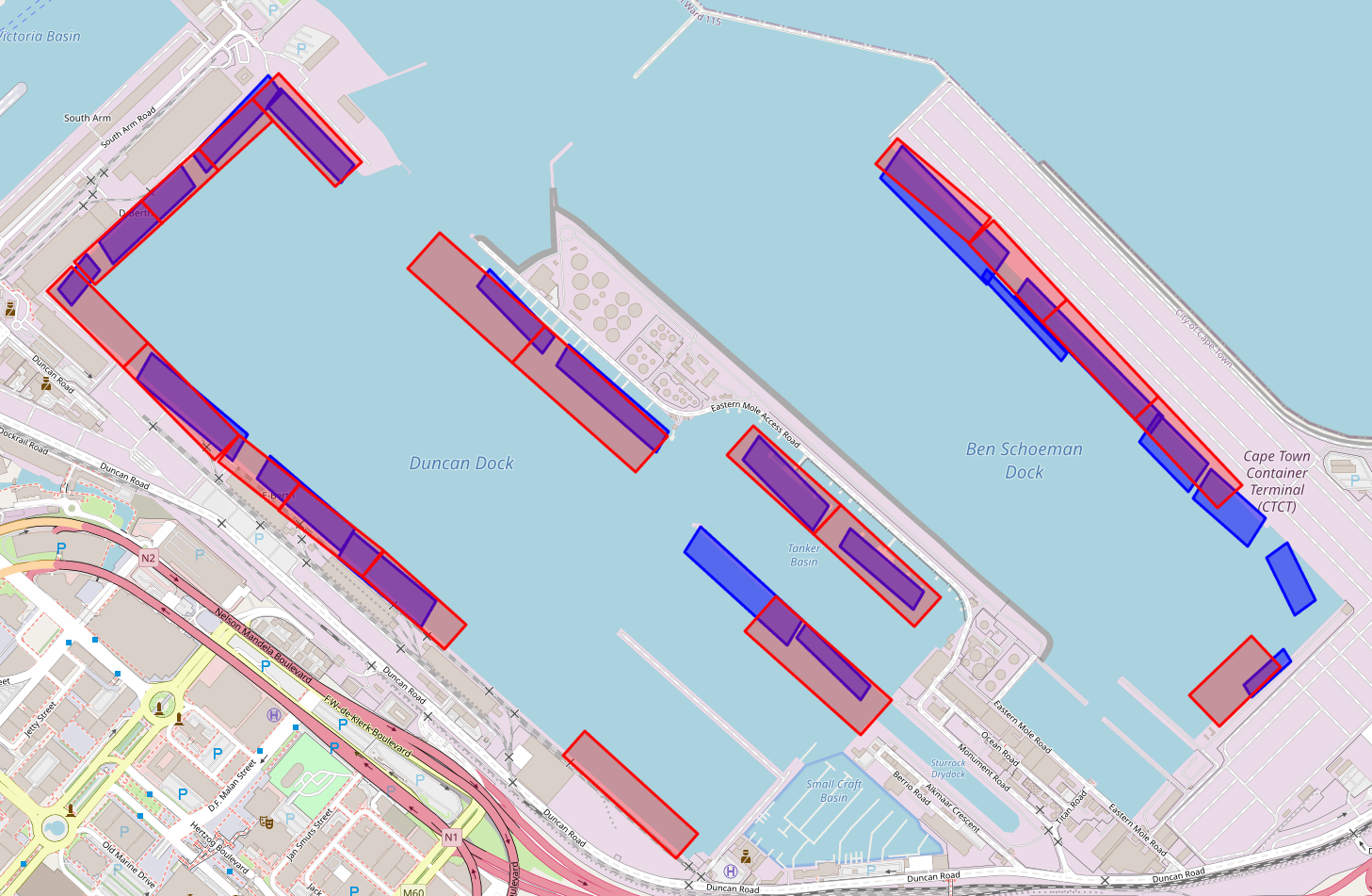 }
    \includegraphics[width=0.32\linewidth]{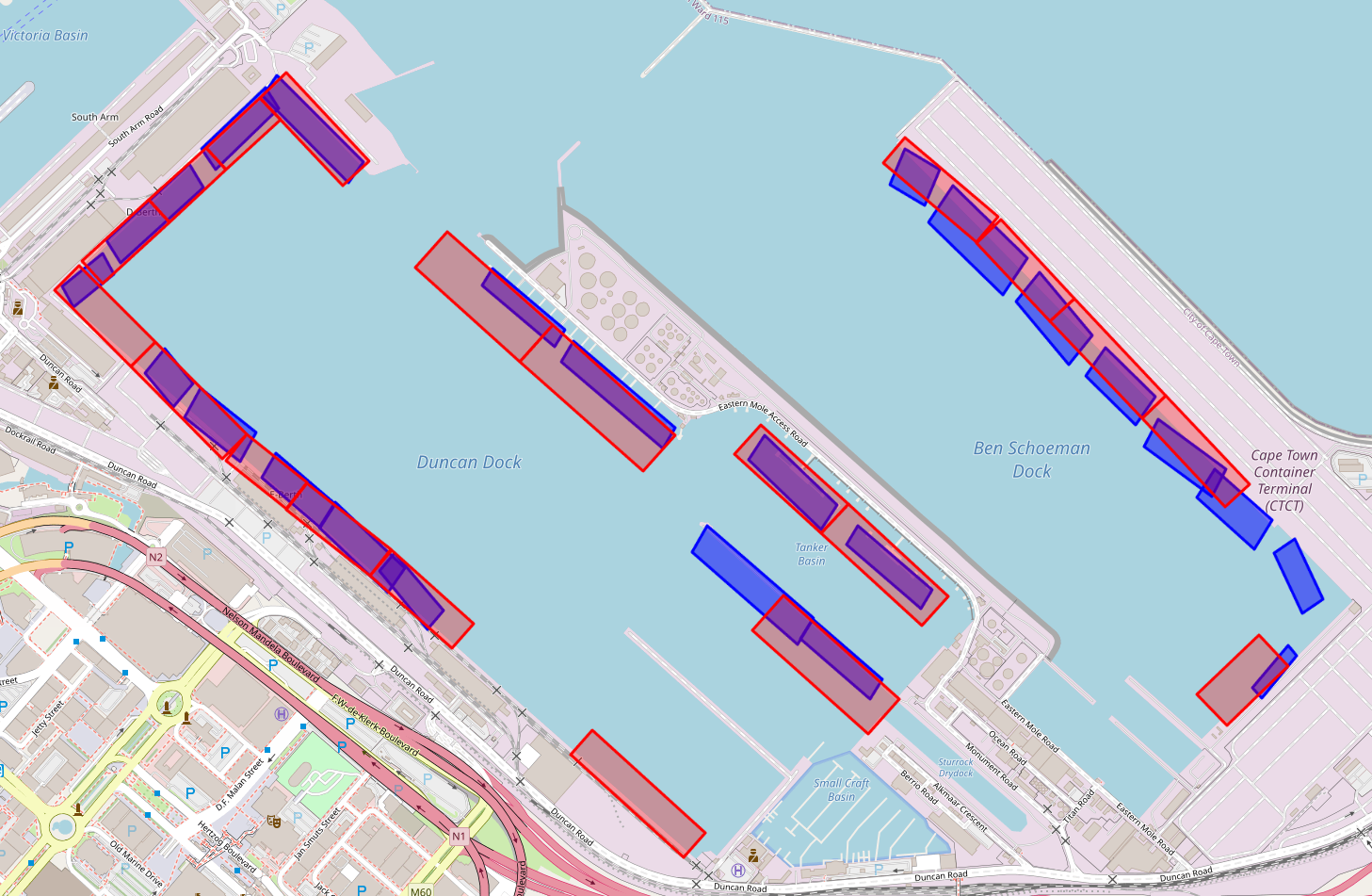 }
    \\
    \includegraphics[width=0.32\linewidth]{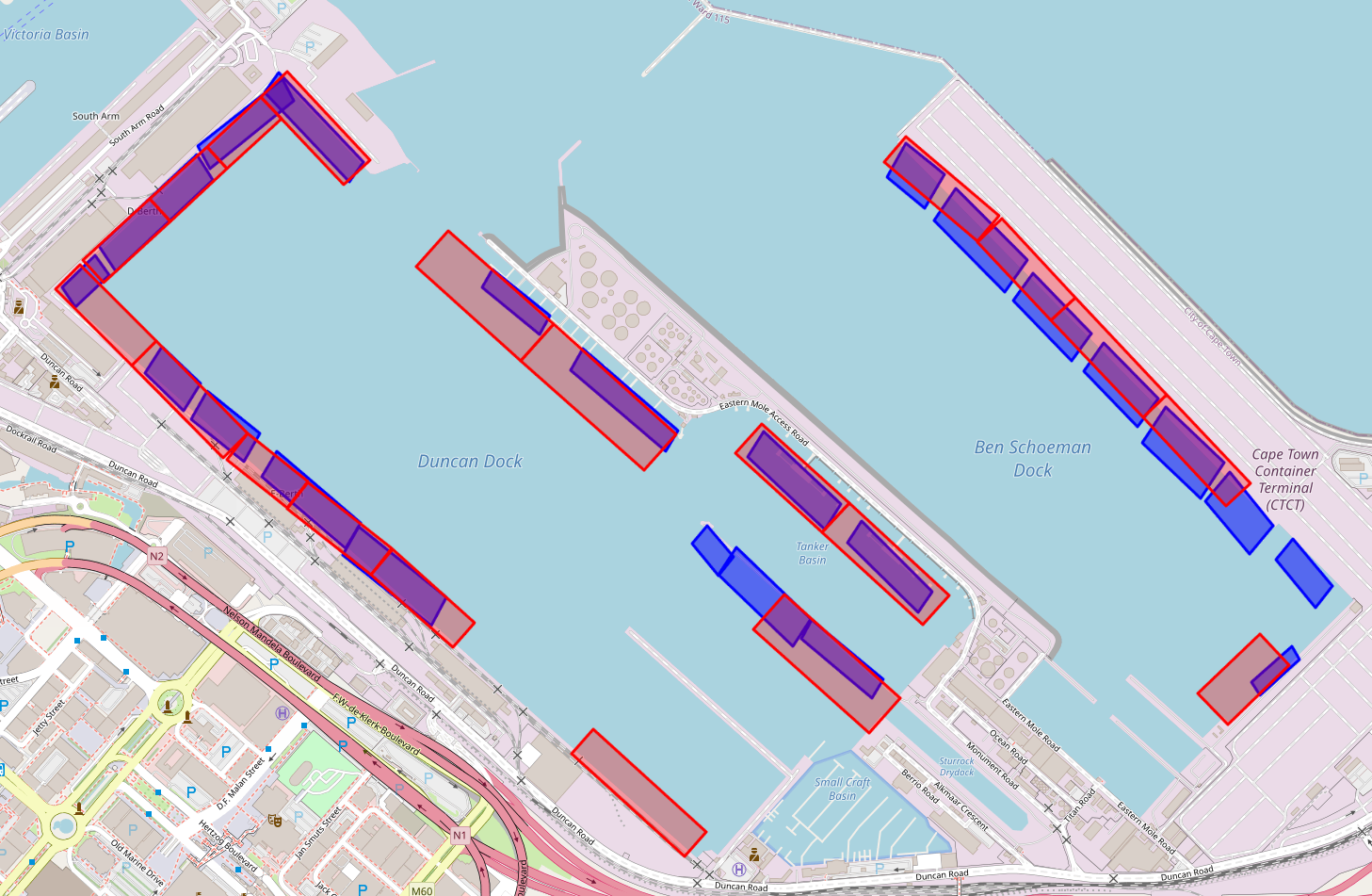}
    \includegraphics[width=0.32\linewidth]{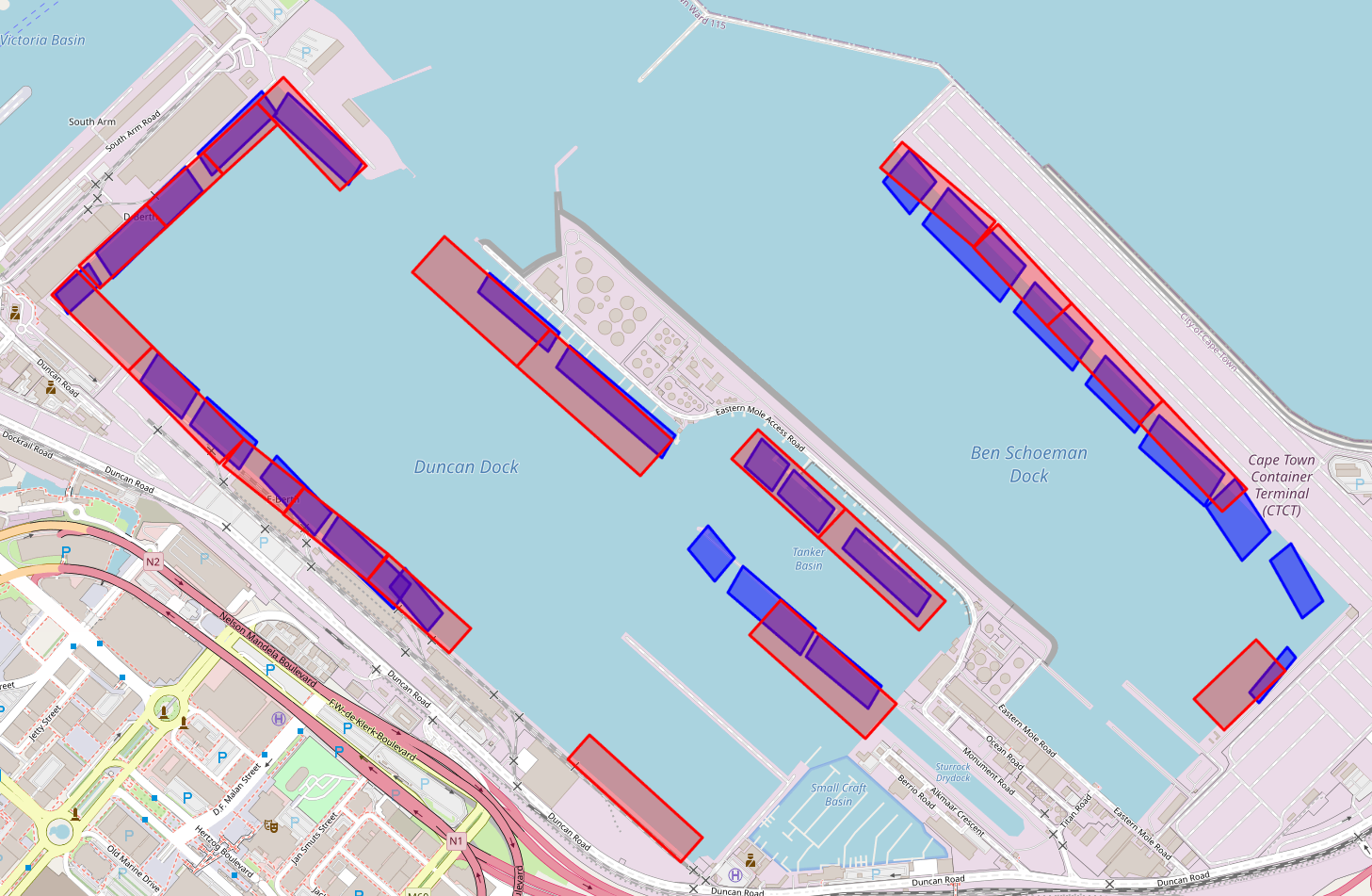}
    \includegraphics[width=0.32\linewidth]{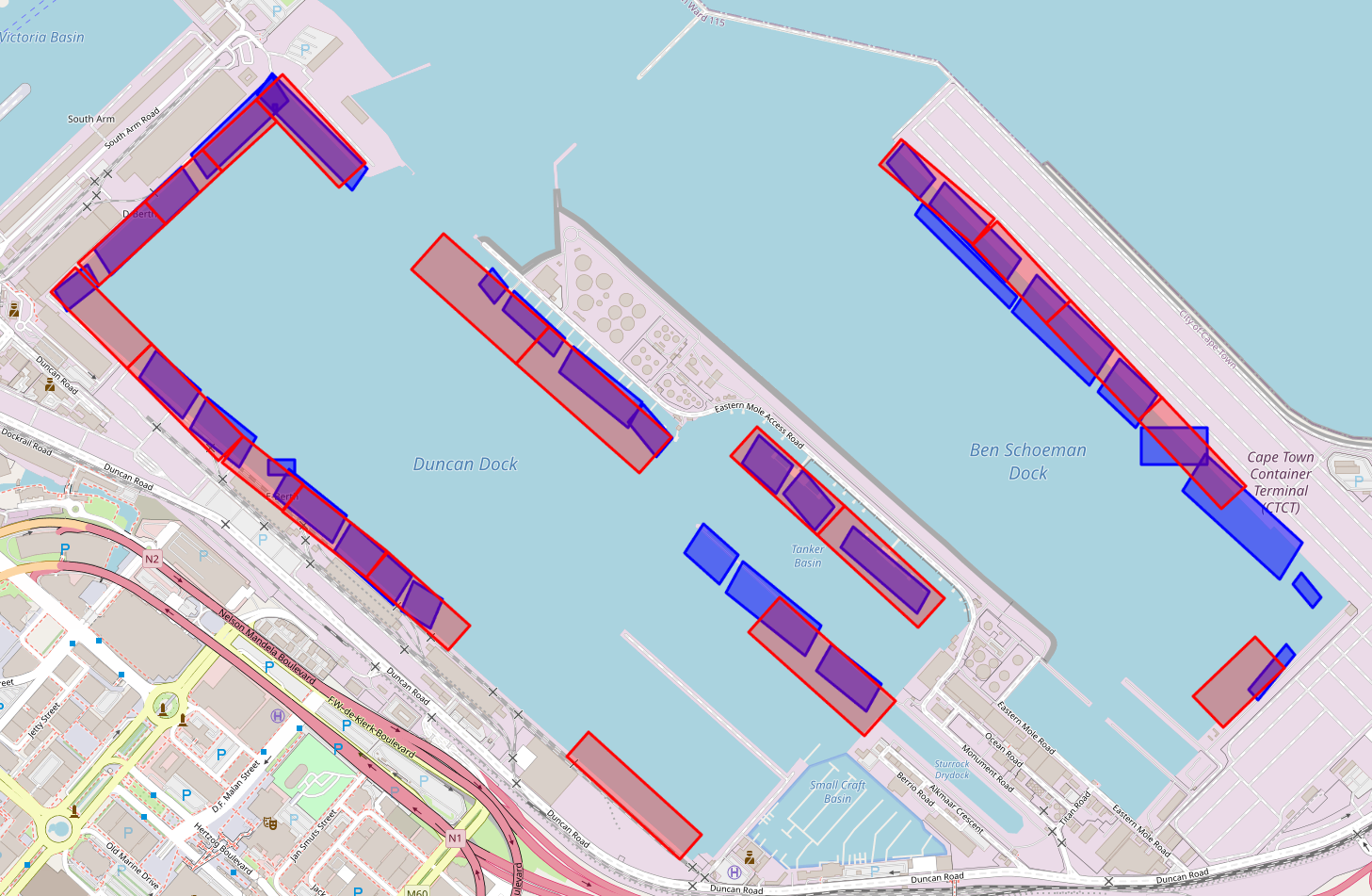}
% \end{adjustwidth}
\caption{Qualitative comparison on the effect of the number of generated points in the port of Cape Town for the proposed geohash-enabled method. From left to right and top to bottom the figures correspond to results of generating $0$, $2$, $5$, $10$, $20$, and $40$ per AIS message.}
\label{fig:gen_ablation}
\end{figure}

Notably, when no additional points are generated (0 points), the Bhattacharyya distance is significantly higher, indicating poorer clustering consistency and underscoring the effectiveness of the proposed spatial augmentation. Qualitative results in Figure~\ref{fig:gen_ablation} corroborate this observation.
% , and similar findings are reported for the non-geohash variant in Appendix B.

Moreover, Table~\ref{tab:ablation_gen} suggests that increasing the number of generated points generally improves clustering consistency. However, the incremental gains diminish as we move toward higher point counts. While adding more points beyond $10$ or $20$ continues to yield marginal improvements, these may not justify the associated computational costs. The qualitative comparison in Figure~\ref{fig:gen_ablation} further supports the above conclusion as minimal differences at higher point densities are observed. 
% Notably, for the non-geohash variant, increasing the number of generated points does not consistently improve performance (see Appendix B for further details). 
Based on these insights, we chose to generate $10$ points during training and $20$ points during evaluation.

% ~\ref{fig:gen_ablation_violin}
% Table~\ref{tab:ablation_gen} shows the Bhattacharyya distance of these results and Figure~\ref{fig:gen_ablation_violin} shows the trends for the choise of number of generated points. 
% Figure~\ref{fig:gen_geo_ablation_qual} illustrates the qualitative evaluation of number of generated points 
% for the Non-Geohash method while Figure~\ref{fig:gen_geo_ablation_qual} 
% for the Geohash method.
% the illustrates that omitting this step in our method leads to the formation of smaller clusters, which frequently exhibit incorrect and unrealistic directions. 
\subsubsection{Interpolation period}
This ablation study evaluates the sensitivity of the proposed method to varying interpolation periods in data preprocessing. Table~\ref{tab:ablation_gen} shows that 15-minute, 30-minute, and 1-hour intervals yield comparable Bhattacharyya distances. However, shorter intervals (15 and 30 minutes) produce qualitatively unnatural berth delineations (see Figure~\ref{fig:inter_ablation_qual}), overly-fragmenting the coastline into too many berths. Increasing to 1 hour preserves stable delineations without unnecessary complexity, balancing both performance and computational cost. Extending to 2 hours reduces computational cost further but introduces a mild performance decline as seen in Table~\ref{tab:ablation_gen}. 
% For the non-geohash variant, similar patterns are observed as discussed in Appendix B. 
Hence, the 1-hour interpolation interval was chosen.

% We tested intervals of 15 minutes, 30 minutes, 1 hour, and 2 hours, and report the Bhattacharyya distances for the Port of Cape Town in Table~\ref{tab:ablation_inter}.

% We carried out an analyses of how much our methods are sensitive to the interpolation period that is used in the preprosessing of the data. Table~\ref{tab:ablation_gen} shows the Bhattacharyya distance of the different interpolation periods that were explored (15 minutes, 30 minutes, 1 hour, and 2 hours). Figure~\ref{fig:inter_ablation_qual} shows the qualitative evaluation of this ablation study, 

% while Figure~\ref{fig:inter_ablation_violin} shows the trend on the difference of different interpolation periods. 
    
% Notably, the Geohash method demonstrates a consistently lower distance in all ports, indicating its effectiveness at the problem at hand.
 %    A more comprehensive analysis of the effect of the amount 
 % of generation of points can be seen in figure \ref{fig:gen_ablation} and the Bhattacharyya distances for this ablation study can be seen in Table~\ref{tab:ablation_gen}.
% \end{itemize}
% \begin{figure}
% % \begin{adjustwidth}{-\extralength}{0cm}
% \centering
% \includegraphics[width=\linewidth]{aug_ablation.drawio (1).png}
% % \end{adjustwidth}
% \caption{Qualitative comparison on the effect of the amount of generated points in the port of Cape Town for the proposed geohash-enabled method. From left to right and top to bottom the figures correspond to results of generating $0$, $2$, $5$, $10$, $20$, and $40$ per AIS message.}
%  \label{fig:gen_ablation}
% \end{figure}  

\section{Discussion}
\label{discussion}
In this work, we presented an unsupervised framework for port berth localization, evaluating two variants—one employing geohash encoding and one without—across a diverse set of 11 ports worldwide. Our results demonstrate a significant advancement over existing approaches in both quantitative and qualitative terms. By comparing our model predictions against satellite imagery and existing berth labels, we highlighted the limitations and inconsistencies in current documentation. As presented, our models were able to identify berths overlooked or inaccurately represented by publicly available labels. Furthermore, the broad evaluation, encompassing ports of varying sizes and operational contexts, attests to the adaptability and generalizability of our approach.

A key factor in the success of our method lies in several novel methodological choices. First, the introduction of a spatial data augmentation strategy, guided by vessel dimensions and headings, enriched the model’s spatial representation and contributed to more coherent berth delineations. Second, our use of a KL-divergence-based score, coupled with Bayesian optimization and the Minimum Description Length (MDL) principle, facilitated a principled and data-driven approach to hyperparameter tuning and model selection. Finally, the post-processing procedure, which transforms the probabilistic outputs of Gaussian Mixture Models (GMMs) into polygonal berth boundaries, enabled a more intuitive and practical interpretation of the results. Each of these components, when combined, provided a robust and flexible framework capable of delivering reliable berth localization results across diverse maritime environments.

Our ablation studies further elucidated the factors influencing berth localization performance. Increasing the observation period (e.g., from a few days to one month) consistently yielded more stable and accurate GMM outputs, confirming the intuitive notion that more AIS messages provide a firmer statistical basis for identifying berths. Likewise, introducing spatial augmentation significantly enhanced clustering consistency, though the marginal gains diminished as the number of generated points increased, guiding us toward a balanced configuration. Setting the interpolation interval to one hour provided balance between computational efficiency and performance. Notably, these adjustments benefited the geohash-enabled variant more consistently than its non-geohash counterpart, underscoring the advantages of encoding spatial information into geohashes. While non-geohash models still improved through augmentation and tuning, their gains were not as pronounced, reinforcing the conclusion that geohash encoding provides a more robust and reliable foundation for data-driven berth localization.

Despite these advances, some limitations remain, offering fertile ground for future investigation. Currently, our approach focuses primarily on cargo and tanker vessels; extending coverage to other vessel types, such as passenger or fishing ships, could broaden the applicability of our method. While the geohash-enabled variant consistently demonstrated superior performance, some of our methodological choices and hyperparameter configurations (e.g.\ interpolation period) were designed to be generally effective across a diverse set of ports, potentially biasing the outcome toward a ``one-size-fits-all'' solution. Future work could tailor these configurations specifically to each variant and port, potentially revealing scenarios where the non-geohash variant might excel or reducing the need for certain preprocessing steps. Additionally, while our method leverages freely available terrestrial AIS data, regions with sparse coverage or data quality issues (e.g.~Port of Ambarli) may pose challenges to the application of our method. 

By bridging the gap between raw AIS data and port berth localization, this work provides a strong foundation for more informed decision-making within maritime logistics and port management. As global data accessibility continues to improve, our framework’s adaptability and scalability position it as a valuable tool for stakeholders aiming to optimize port operations. Ultimately, advances in unsupervised berth localization can pave the way for more agile, data-driven maritime strategies and more resilient global trade networks.

\section{Declaration of generative AI and AI-assisted technologies in the writing process}
During the preparation of this manuscript, the authors used ChatGPT to improve the phrasing of various text segments. After using this tool/service, the author(s) reviewed and edited the content as needed and take(s) full responsibility for the content of the publication.
\section{Funding}
This work was supported through funding by the Republic of Cyprus through the Research and Innovation Foundation project with grant number CODEVELOP-GT/0322/0096 (ADAPTATION), and the EU H2020 Research and Innovation Programmes under Grant Agreement No. 857586 (CMMI-MaRITeC-X) and No. 101158669 (AXOLOTL). Views and opinions expressed are however those of the authors only and do not necessarily reflect those of the funding authorities. 
% Neither the European Union nor the granting authority can be held responsible for them.
\section{Acknowledgments}
Our work would not be possible without the support of our colleague Charalambos Rotsides who was instrumental in the efficient collection, processing, and storage of the AISStream data.
\appendix
% \section{Appendix}
% \renewcommand\thefigure{\thesection.\arabic{Figure}}  
% \renewcommand\thefigure{\thesection.\arabic{Table}}  
\counterwithin{figure}{section}
\counterwithin{table}{section}
\section{Data preprocessing}
\label{Data preprocessing appendix}

\begin{table}[!htbp]
% \resizebox{\textwidth}{!}
\caption{The table illustrates the quantity of AIS messages remaining after each step in the preprocessing sequence for the one-month period. This sequence is arranged in a left-to-right order in the table, reflecting the chronological progression of the preprocessing stages. Specifically, the data undergoes the following steps: first, it is filtered by Area of Interest (AOI), then by maximum speed. Next, records with a heading value of ``$511$" are filtered out. The data is then split into two parts, after which interpolation is applied. Finally, records with a maximum heading difference between subsequent entries are filtered out.}
\label{tab:1month}
\scalebox{0.75}{
\begin{tabular}{ccccccccccc}
\toprule
\textbf{Port}        & \textbf{AIS data} & \textbf{AOI} & \textbf{Speed} & \textbf{511 heading} & \multicolumn{2}{c}{\textbf{Split}} & \multicolumn{2}{c}{\textbf{Interpolation}} & \multicolumn{2}{c}{\textbf{Heading}} \\ \midrule
\textbf{Singapore}   & 609975            & 119466       & 103624         & 95369                & 47870            & 47499           & 25561                & 25414               & 23559             & 23483            \\
\textbf{Busan}       & 120752            & 90363        & 76037          & 54491                & 27477            & 27014           & 14753                & 14712               & 10636             & 10124            \\
\textbf{Antwerp}     & 676896            & 322840       & 281909         & 104479               & 52514            & 51965           & 27950                & 27666               & 25723             & 25539            \\
\textbf{Los Angeles} & 42055             & 11249        & 10954          & 10954                & 5575             & 5379            & 2822                 & 2743                & 2758              & 2684             \\
\textbf{Southampton} & 24998             & 8383         & 7382           & 7243                 & 3905             & 3338            & 2031                 & 1746                & 1872              & 1651             \\
\textbf{Auckland}    & 5618              & 5587         & 5587           & 5184                 & 2919             & 2265            & 1484                 & 1174                & 1420              & 1115             \\
\textbf{Livorno}     & 19700             & 14476        & 13382          & 13127                & 6690             & 6437            & 3455                 & 3313                & 3304              & 3148             \\
\textbf{Cape Town}   & 17863             & 11030        & 10444          & 10398                & 5398             & 5000            & 2757                 & 2535                & 2655              & 2470             \\
\textbf{Gdansk}      & 72020             & 38192        & 37311          & 32314                & 16480            & 15834           & 8404                 & 8079                & 8230              & 7910             \\
\textbf{Limassol}    & 10049             & 2131         & 1899           & 1690                 & 874              & 816             & 464                  & 441                 & 404               & 402              \\
\textbf{Algeciras}   & 76008             & 15368        & 12335          & 11971                & 6099             & 5872            & 4173                 & 3957                & 3058              & 2714             \\
\bottomrule
\end{tabular}
}
\end{table}

Table~\ref{tab:1month} details the number of AIS messages after each prepossessing step over the POI and across all ports under investigation. Initially, the data are refined by selecting only records with a speed below a maximum threshold. Next, AIS messages with missing dimension information are excluded - however, all AIS messages in our experiments were complete in that sense. Records with a heading of ``$511$'', indicating unavailable heading data, are then discarded. The data are then divided into two approximately equal parts (as described in the main text). In the interpolation step, data for each vessel is interpolated on an hourly basis. Finally, any records with significant heading changes (herein defined as larger than $10$ degrees) are removed from these splits.

\section{Optimal hyperparameter values}
\label{Optimal hyperparameter values appendix}
The optimal hyperparameter values for both DBSCAN and GMM, as determined through our model selection process, are presented in Table~\ref{tab:hyperparam_3d_1month} for all ports of interest, encompassing both the geohash and non-geohash variants of the proposed method.
\begin{table}[!htbp]
\caption{Optimal hyperparameters for both the Geohash and Non-Geohash methods over a one-month period.}
\scalebox{0.75}{
\begin{tabular}{clccccccc}
\hline
\textbf{Period: 1 month} &                      & \multicolumn{3}{c}{\textbf{Non-Geohash}}                         & \textbf{} & \multicolumn{3}{c}{\textbf{Geohash}}                             \\ \hline
\textbf{Port}            &                      & \textbf{Epsilon (m)} & \textbf{MinPoints} & \textbf{nComponents} &           & \textbf{Epsilon (m)} & \textbf{MinPoints} & \textbf{nComponents} \\ \hline
\textbf{Algeciras}       &                      & 33.509               & 2                  & 49                   &           & 35.215               & 2                  & 49                   \\
\textbf{Aukland}         &                      & 37.445               & 2                  & 27                   &           & 34.841               & 2                  & 21                   \\
\textbf{Antwerp}         & \multicolumn{1}{c}{} & 14.630               & 2                  & 228                  &           & 53.296               & 3                  & 228                  \\
\textbf{Busan}           &                      & 34.687               & 2                  & 49                   &           & 5.630                & 7                  & 48                   \\
\textbf{Cape Town}       &                      & 55.004               & 9                  & 42                   &           & 35.454               & 13                 & 23                   \\
\textbf{Gdansk}          &                      & 52.742               & 5                  & 49                   &           & 22.829               & 4                  & 48                   \\
\textbf{Limassol}        &                      & 17.319               & 3                  & 10                   &           & 37.060               & 2                  & 10                   \\
\textbf{Livorno}         &                      & 27.159               & 2                  & 49                   &           & 15.778               & 5                  & 41                   \\
\textbf{Los Angeles}     &                      & 19.797               & 5                  & 49                   &           & 42.313               & 14                 & 40                   \\
\textbf{Singapore}       &                      & 49.436               & 2                  & 228                  &           & 35.830               & 2                  & 226                  \\
\textbf{Southampton}     &                      & 26.180               & 5                  & 48                   &           & 12.750               & 4                  & 32                   \\ \hline
\end{tabular}

}

\label{tab:hyperparam_3d_1month}
\end{table}

\section{Further ablation results}
\label{Further ablation results appendix}

\subsection{Period of interest (POI)}
\label{Period of interest (POI) appendix}
\begin{table}[!htbp]
\centering
\caption{Ablation study results of the non-geohash variant across the ports showing the effect of the period of interest on performance (Bhattacharyya distance). Left and right columns represent different sets of ports.}
\scalebox{0.80}{
\begin{tabular}{lcccc|lcccc}
\hline
\textbf{Port}     & \textbf{3 days} & \textbf{1 week} & \textbf{2 weeks} & \textbf{1 month} & \textbf{Port}     & \textbf{3 days} & \textbf{1 week} & \textbf{2 weeks} & \textbf{1 month} \\ \hline
                  &                 &                 &                  &                  & Gdansk           & 3.649           & 2.121           & 1.656            & \textbf{1.237}  \\
Algeciras         & 3.677           & 2.302           & 1.661            & \textbf{0.882}  & Limassol         & 1.277           & 1.135           & \textbf{0.939}   & 1.020           \\
Antwerp           & 2.312           & 1.738           & 1.462            & \textbf{0.800}  & Livorno          & 1.786           & 1.467           & 1.444            & \textbf{0.815}  \\
Auckland          & 1.944           & 1.492           & 1.547            & \textbf{1.145}  & Los Angeles      & 6.885           & 1.821           & 1.236            & \textbf{1.235}  \\
Busan             & 1.530           & 1.022           & 0.954            & \textbf{0.950}  & Singapore        & 1.494           & 1.184           & 0.981            & \textbf{0.734}  \\
Cape Town         & 2.160           & 1.386           & 1.329            & \textbf{0.985}  & Southampton      & 2.369           & 1.504           & \textbf{1.097}   & 1.288           \\
\hline
\end{tabular}
}

\label{a_tab:combined_geohash_results}

\end{table}

\begin{figure}[!htbp]
\centering
    \includegraphics[width=0.42\linewidth]{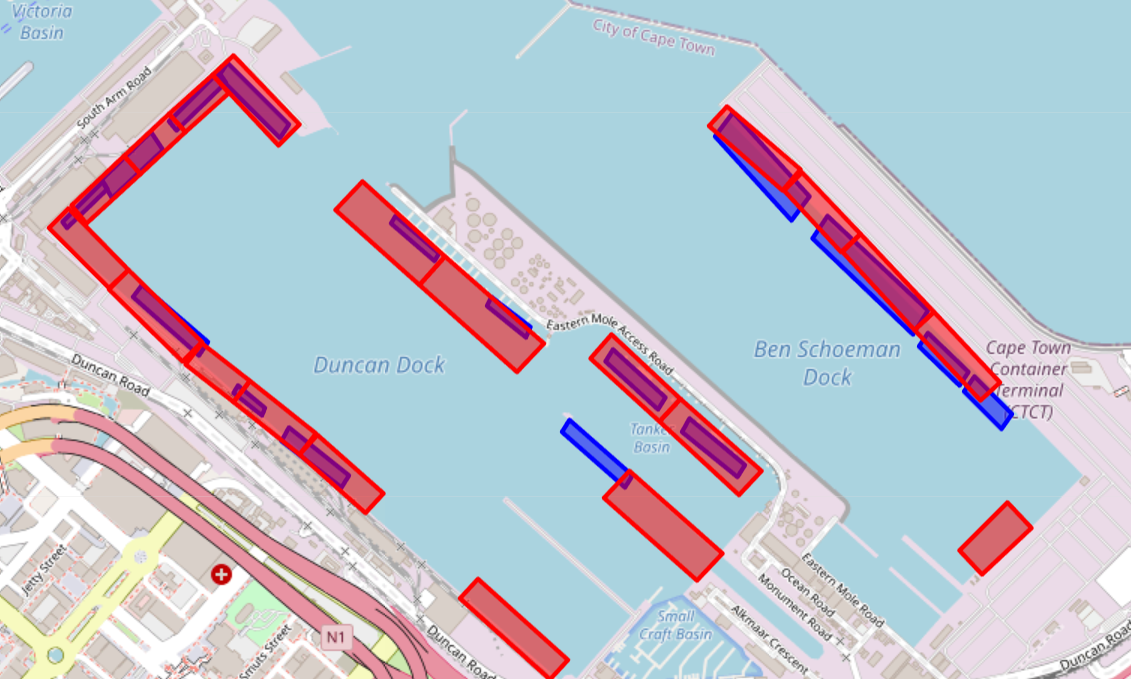}
    \includegraphics[width=0.42\linewidth]{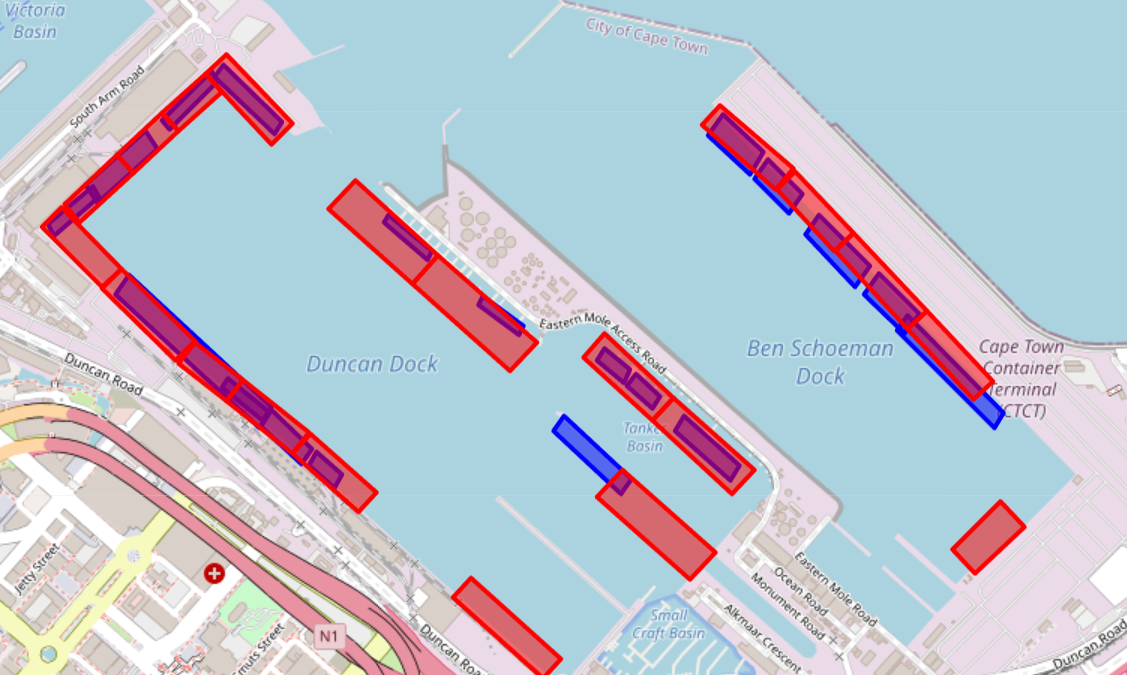}\\
    \includegraphics[width=0.42\linewidth]{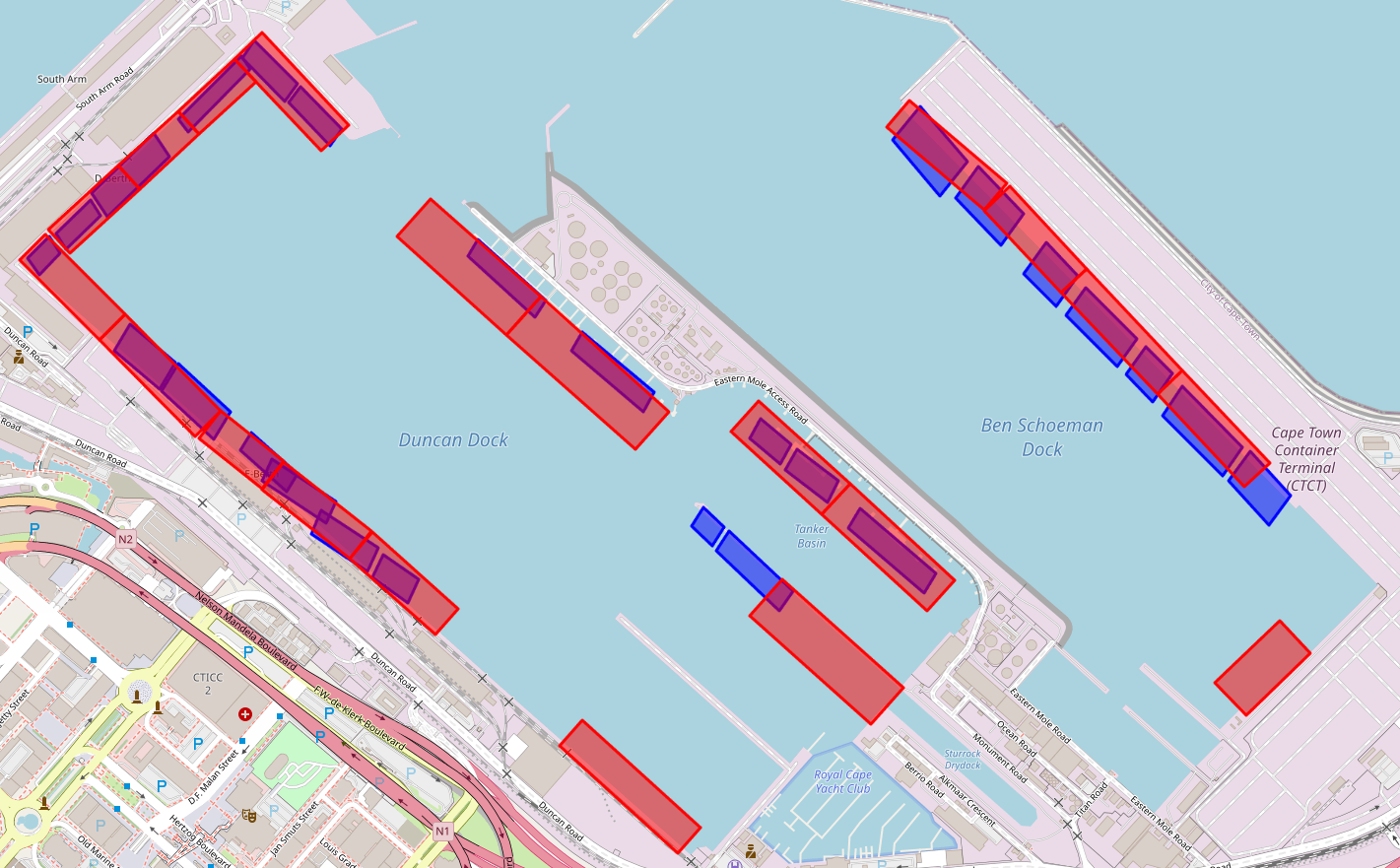}
    \includegraphics[width=0.42\linewidth]{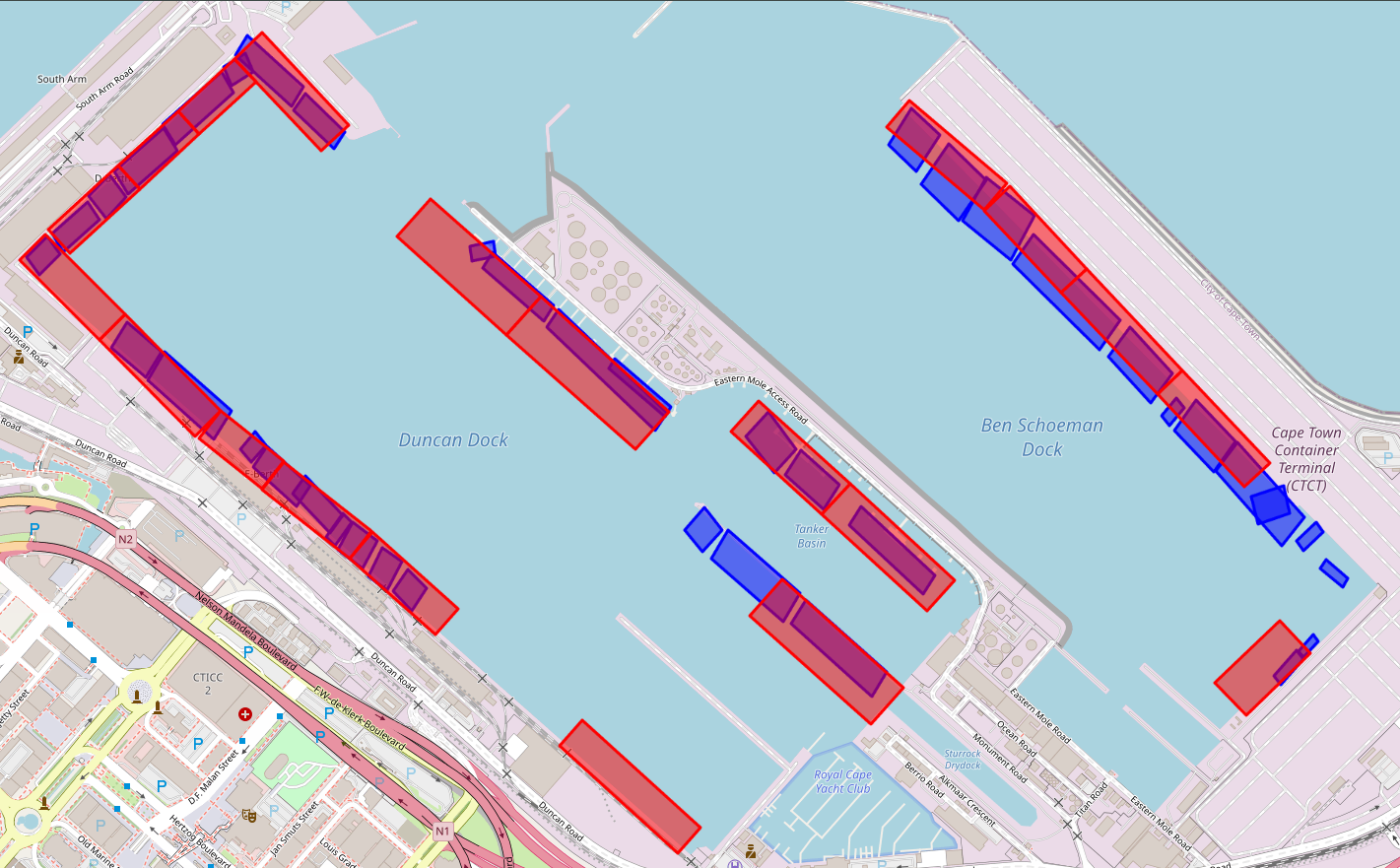}
    % \\
    % \includegraphics[width=0.24\textwidth]{geo_cape_3d (1).png}
    % \includegraphics[width=0.24\textwidth]{geo_cape_1w (1).png}
    % \includegraphics[width=0.24\textwidth]{geo_cape_2w (1).png}
    % \includegraphics[width=0.24\textwidth]{geo_cape_1m (1).png}
    
\caption{Qualitative comparison of the proposed non-geohash variant with increasing period of interest for the port of Cape Town. From left to right and top to bottom  the period is 3 days, 1 week, 2 weeks, and 1 month.}
\label{a_fig:cape}
\end{figure} 
As shown in Table~\ref{a_tab:combined_geohash_results}, increasing the POI duration generally improves GMM consistency across the two data splits, although exceptions exist. Notably, in Limassol and Southampton, the 2-week interval slightly outperforms the 1-month interval, with statistical significance only at Southampton. A qualitative comparison for the Port of Cape Town (Figure~\ref{a_fig:cape}) reveals that while shorter durations (3 days and 1 week) result in sparse berth localization, the difference between 2 weeks and 1 month is less pronounced than in the geohash variant. Nevertheless, given the overall trend and performance, choosing the 1-month POI remains the more favorable option in the non-geohash setting.
\subsection{Number of points generated during spatial augmentation}
\label{Number of points generated during spatial augmentation appendix}

\begin{table}[ht]
% \resizebox{\textwidth}{!}
\centering
\caption{Ablation study results for the Port of Cape Town, illustrating (first two columns) the impact of spatial augmentation and the sensitivity to varying numbers of generated points, and (last two columns) the impact
of different interpolation times in the proposed non-geohash method. Performance is measured using the Bhattacharyya distance.}
\scalebox{0.9}{
\begin{tabular}{cccccc}
\hline
\textbf{\# generated points}      & 
\multicolumn{1}{l}{\textbf{Mean}} & \multicolumn{1}{l}{\textbf{Interpolation period}} & \multicolumn{1}{l}{\textbf{Mean}} \\ \cline{1-4}
\textbf{0 points}  &                      
3.254                             &                                   &                                
\\
\textbf{2 points}  &                      
0.821                            & 15 minutes                                  & 0.818                                                                      \\     
\textbf{5 points}  &                       
0.821                             & 30 minutes                                  & 0.834                                                 
                    
\\
\textbf{10 points} &   
0.825                             & 1 hour                                  & 0.837            &                                              \\
\textbf{20 points} &    
0.820                             &   2 hours                                 & 0.848                                                                     \\
\textbf{40 points} &                       
0.829                             &                                   &                                                             
\\ \hline
\end{tabular}

}

\label{a_tab:ablation_gen}
\end{table}

% This ablation study examines how varying the number of generated points per AIS message affects berth boundary precision and clustering consistency for the non-geohash variant. 
\begin{figure}[!htbp]
% \begin{adjustwidth}{-\extralength}{0cm}
\centering
\includegraphics[width=0.80 \linewidth]{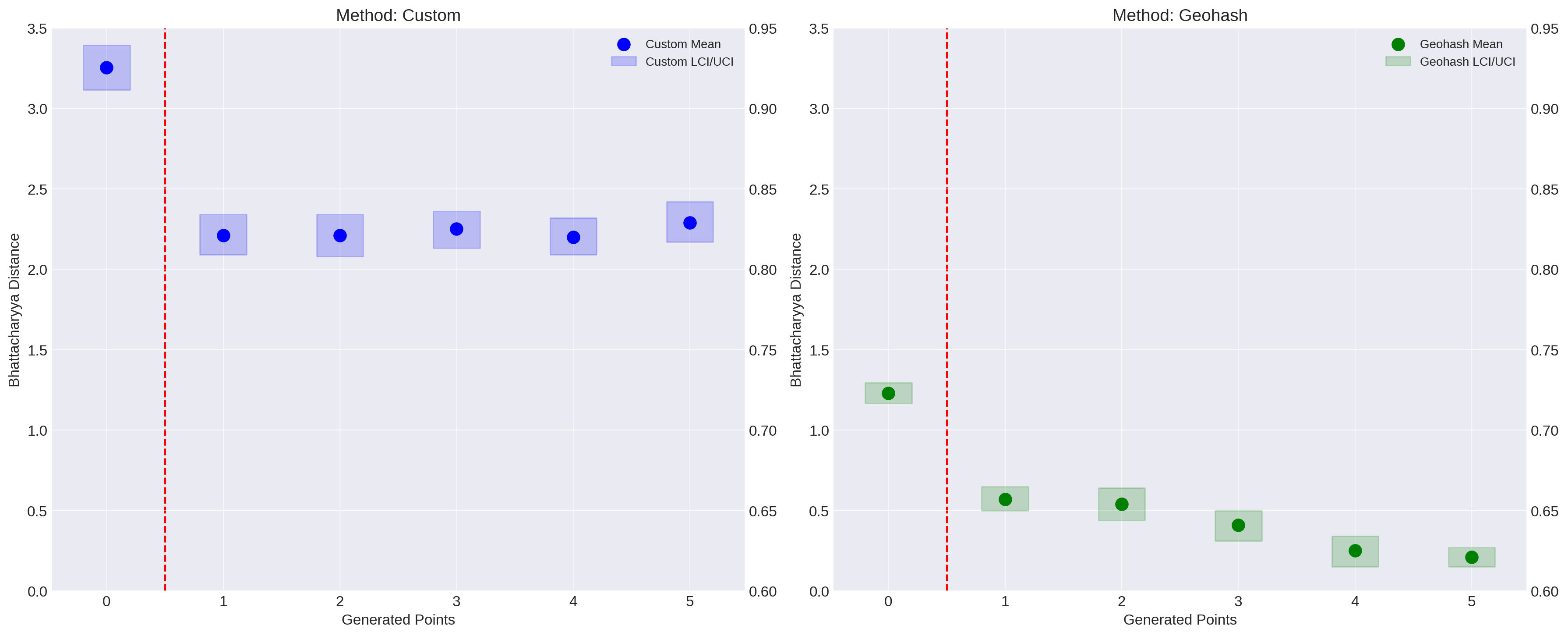}\\
% \end{adjustwidth}
\includegraphics[width=0.80\linewidth]{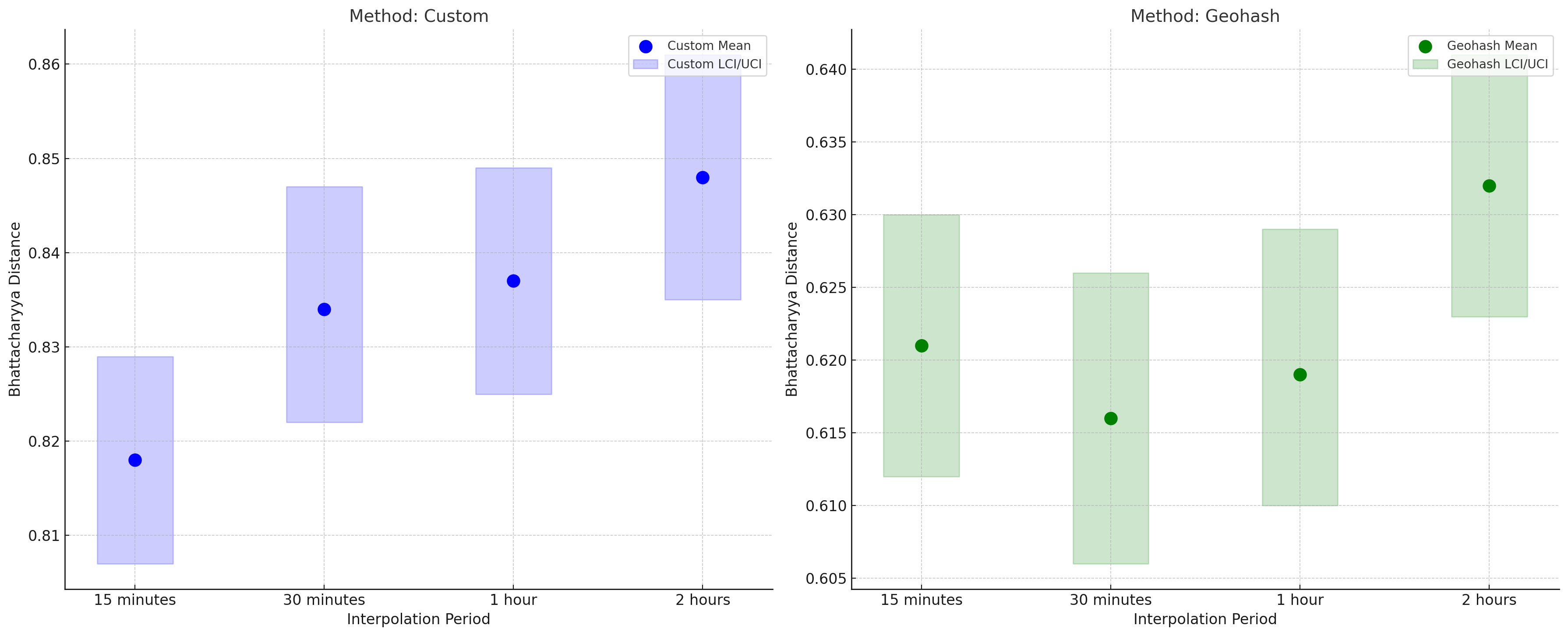}
\caption{Bhattacharyya distances for the generation of points ablation study (top) and the interpolation periods ablation study (bottom). The y-axis for the 0 points generated (top) is in a different range from the rest of the points for better visualization.}
 \label{a_fig:gen_ablation_violin}
\end{figure} 

\begin{figure}[!htbp]
% \begin{adjustwidth}{-\extralength}{0cm}
\centering
    \includegraphics[width=0.3\linewidth]{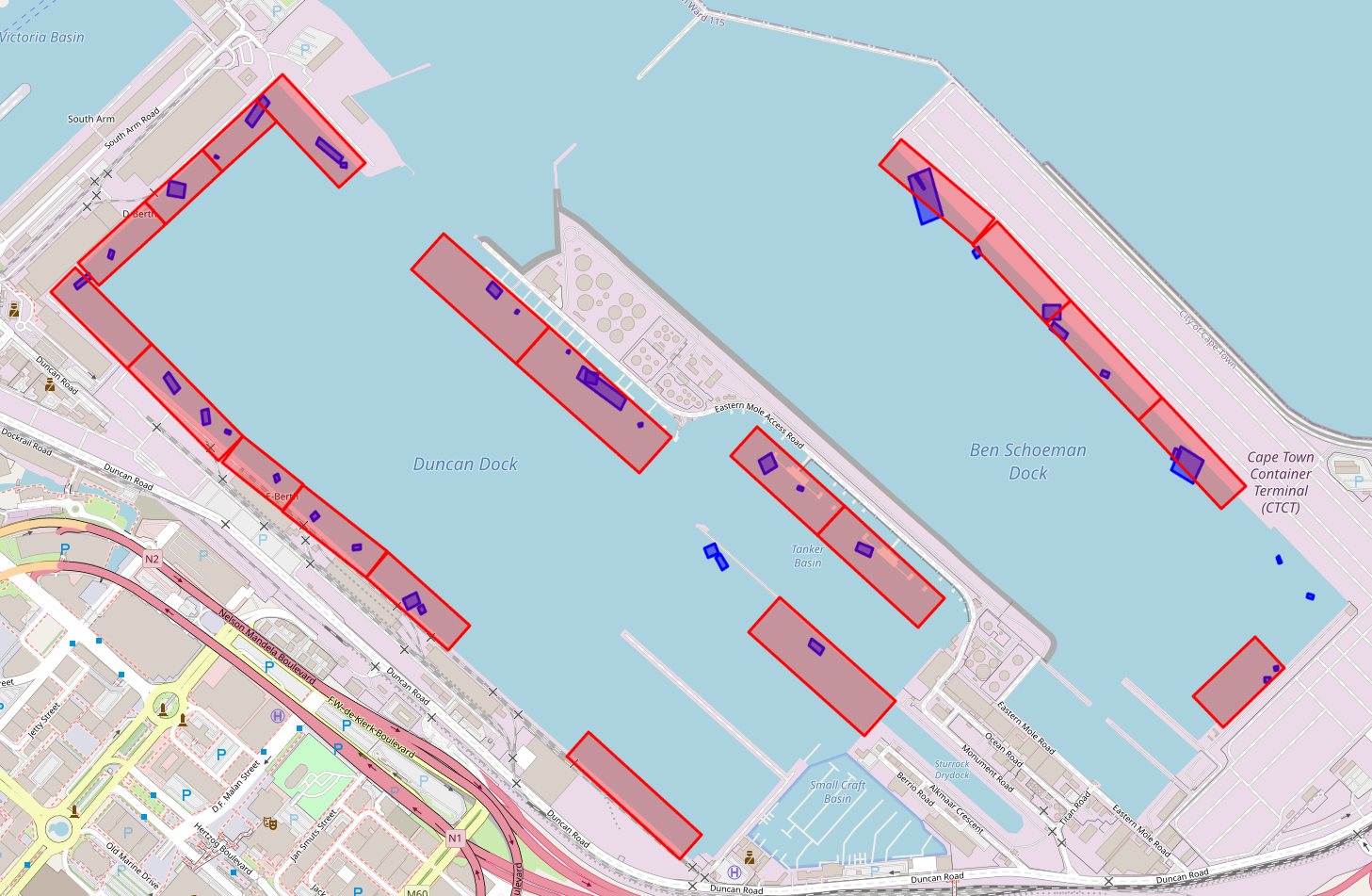 }
    \includegraphics[width=0.3\linewidth]{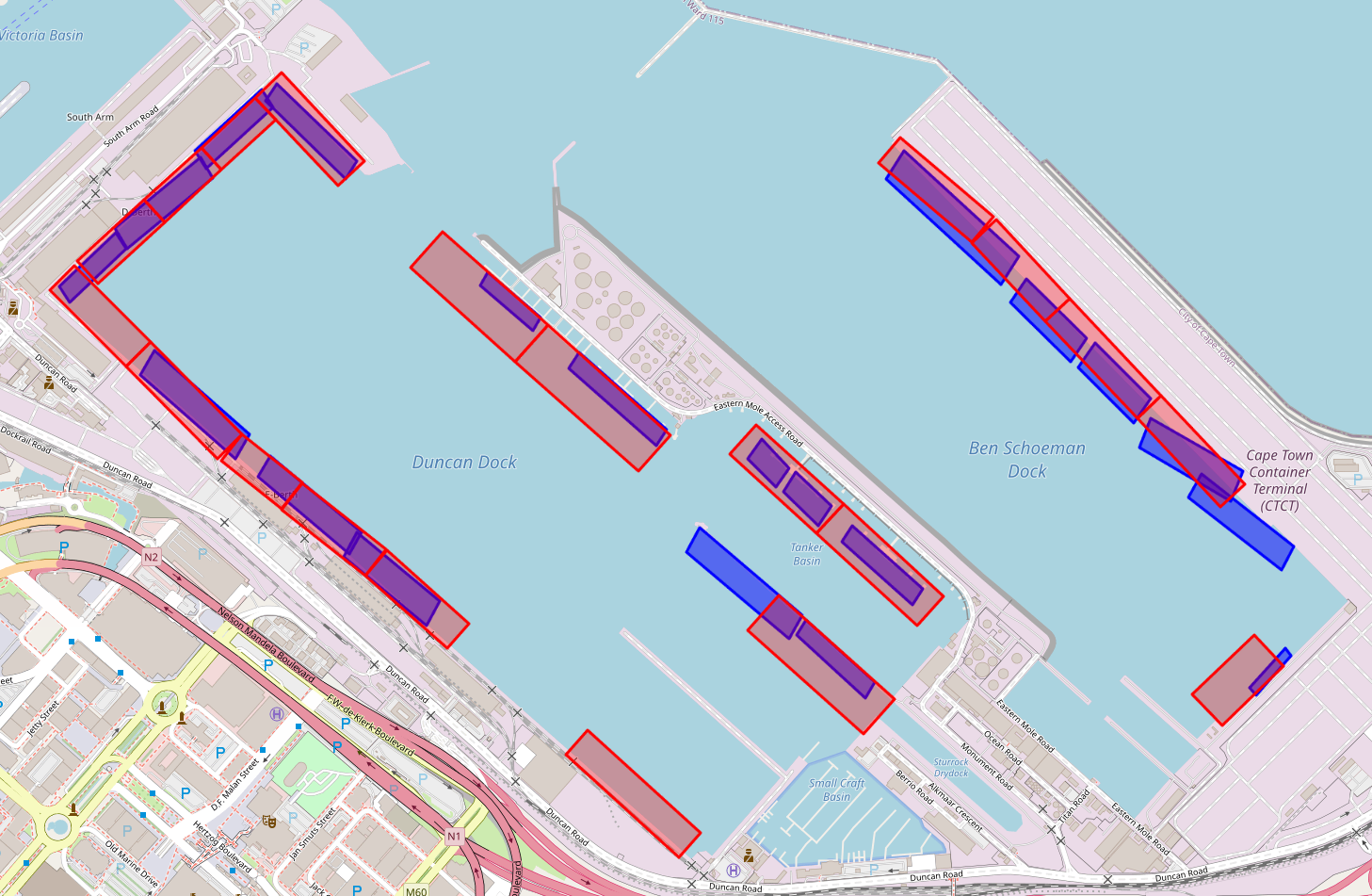 }
    \includegraphics[width=0.3\linewidth]{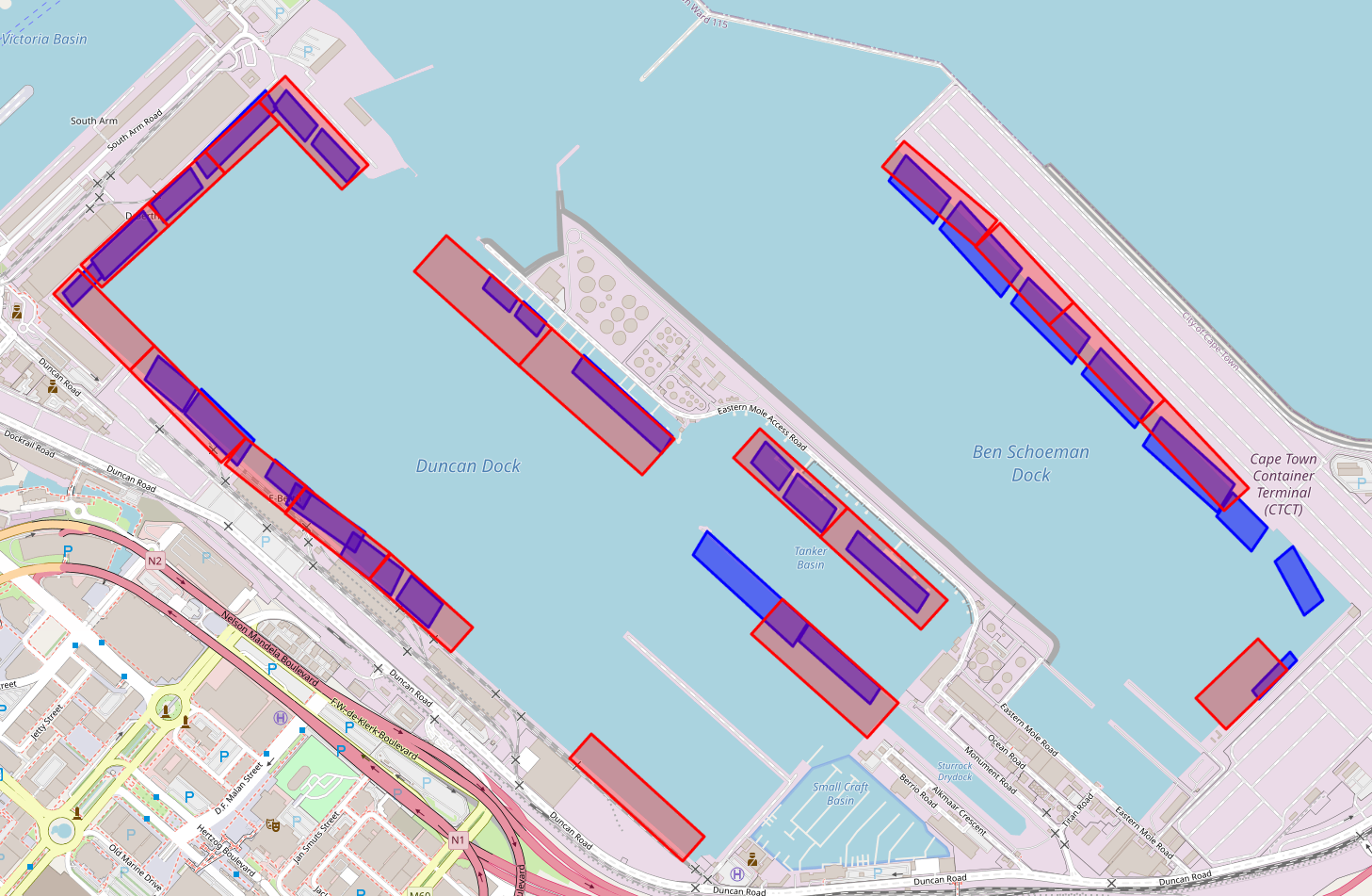 }
    \\
    \includegraphics[width=0.3\linewidth]{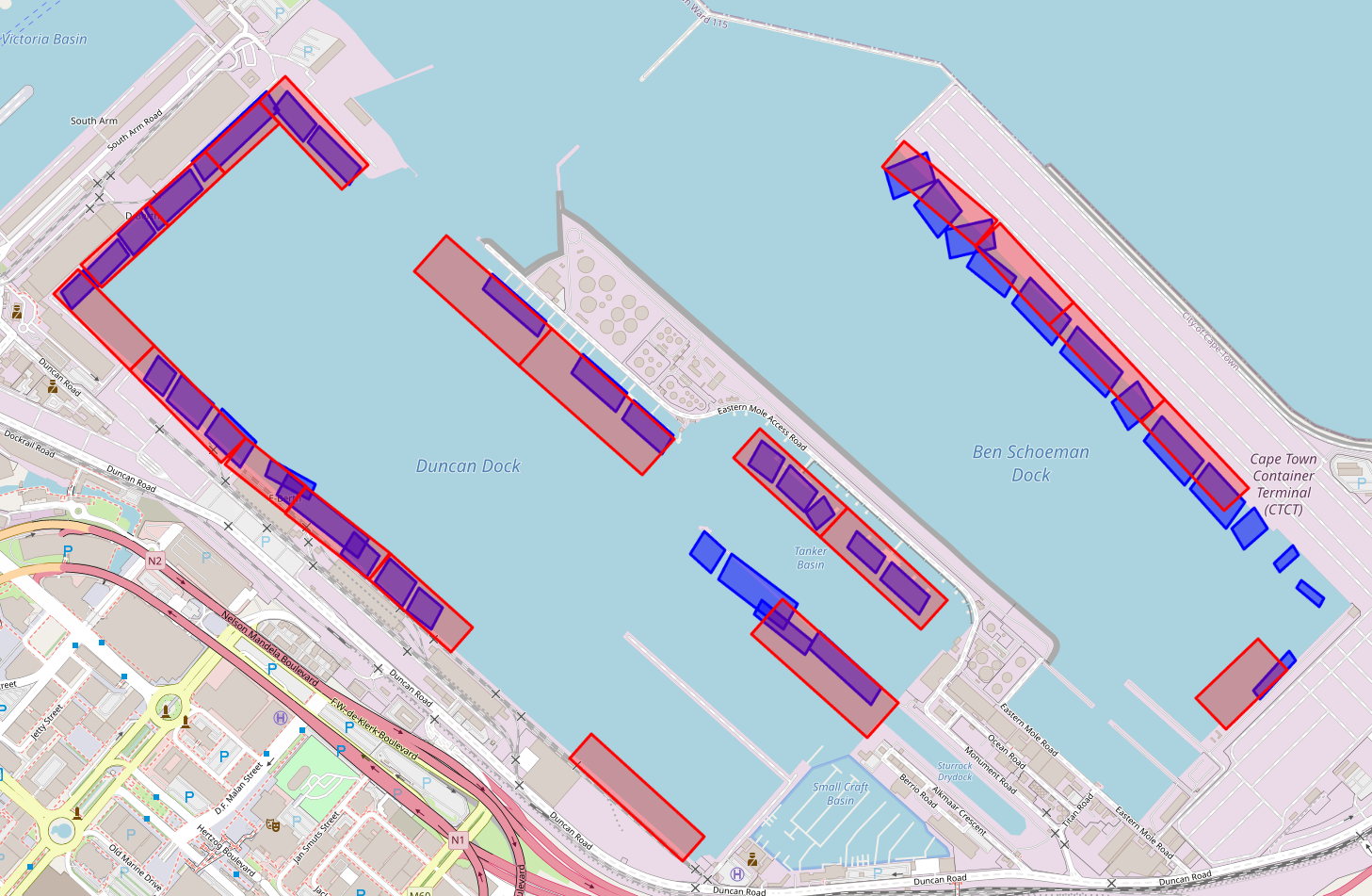}
    \includegraphics[width=0.3\linewidth]{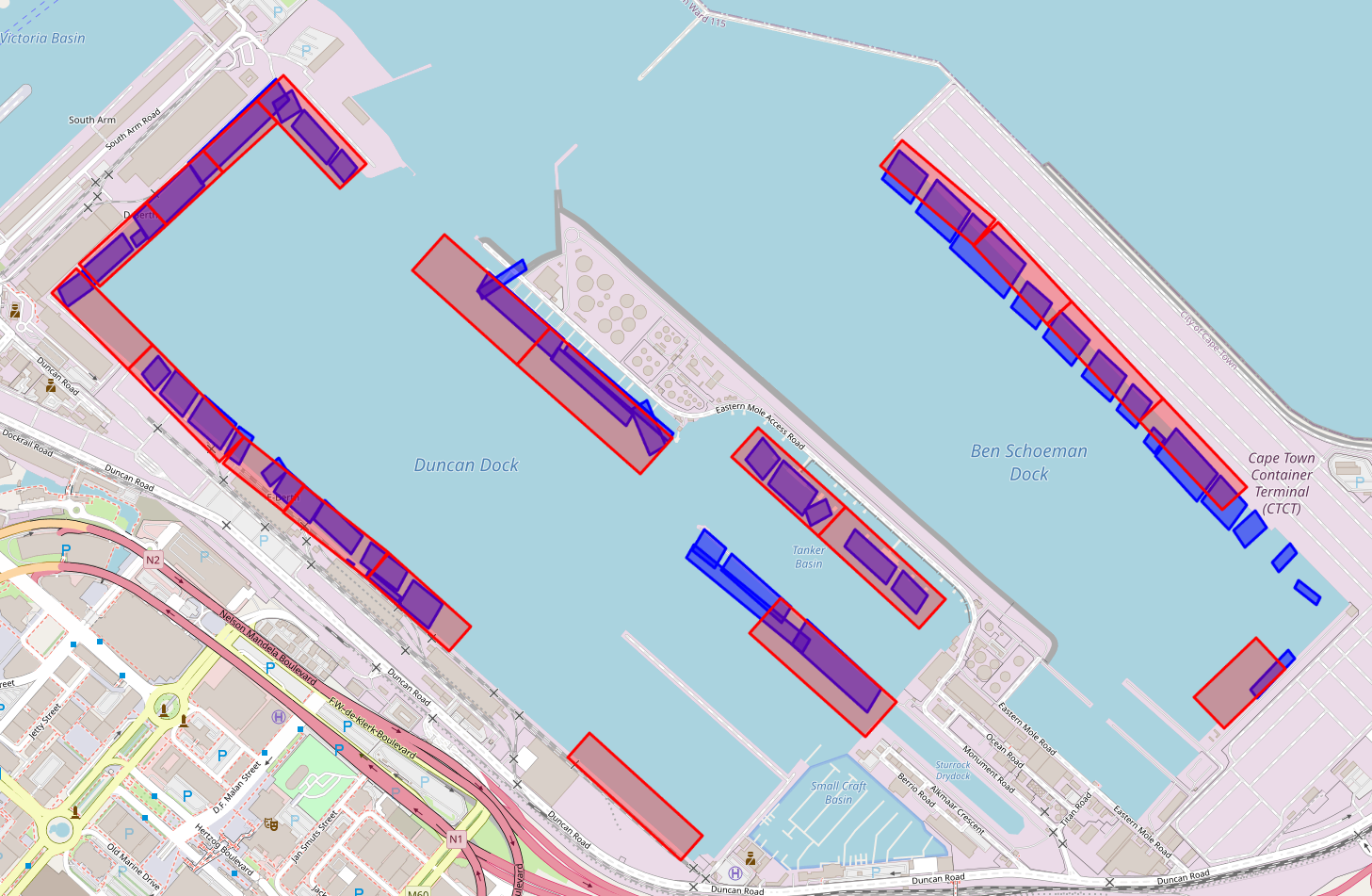}
    \includegraphics[width=0.3\linewidth]{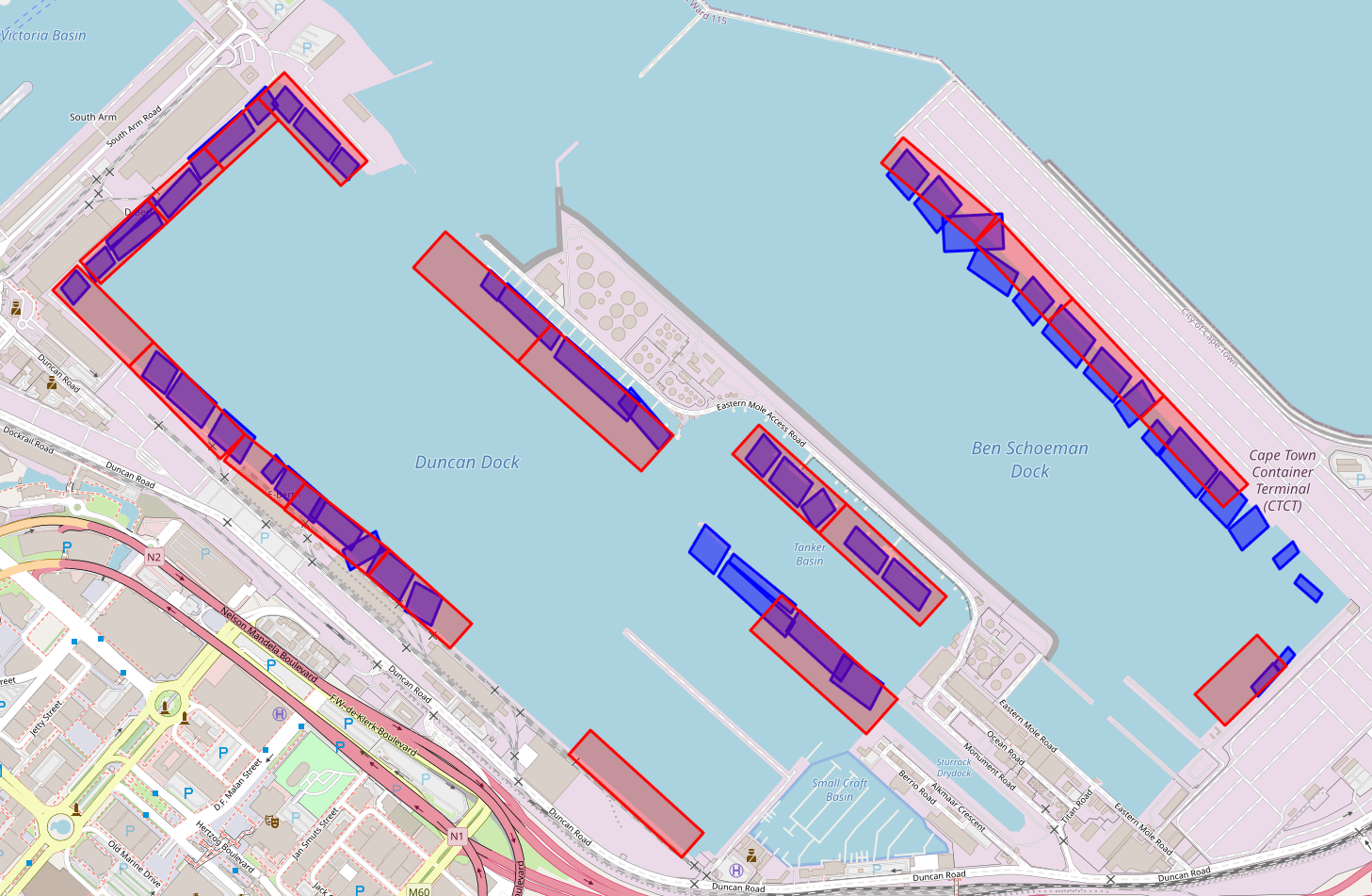}
    
% \end{adjustwidth}
\caption{Qualitative comparison on the effect of the number of generated points in the port of Cape Town for the proposed non-geohash variant. From left to right and top to bottom the figures correspond to results of generating 0, 2, 5, 10, 20, and 40 per AIS
message.}
\label{a_fig:gen_custom_ablation_qual}
\end{figure}
As shown in Table~\ref{a_tab:ablation_gen}, introducing any number of additional points substantially improves clustering consistency compared to having no additional points, confirming the effectiveness of spatial augmentation. However, unlike the geohash-enabled variant, increasing the number of generated points beyond a minimal threshold does not yield clear incremental gains (see Figure~\ref{a_fig:gen_ablation_violin}). Qualitative results in Figure~\ref{a_fig:gen_custom_ablation_qual} further support this: higher point counts neither improve berth delineations nor provide clearer boundaries, and instead risk identifying too many berths (overfitting). Given the lack of quantitative evidence to favor a different approach, we adopt the same configuration as in the geohash variant for consistency.

\subsection{Interpolation period}
\label{Interpolation period}

\begin{figure}[!htbp]
% \begin{adjustwidth}{-\extralength}{0cm}
\centering
    \includegraphics[width=0.425\linewidth]{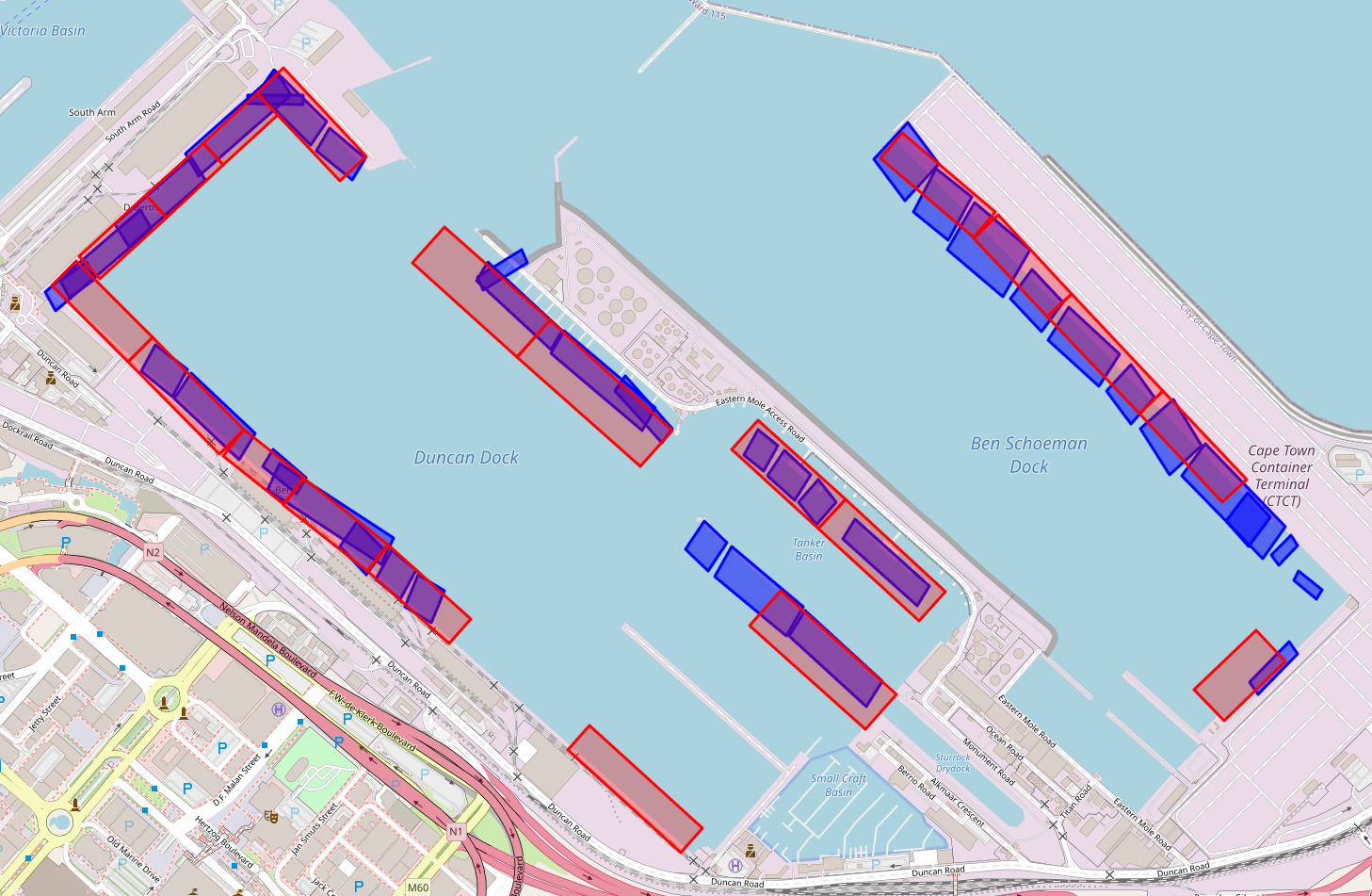}
    \includegraphics[width=0.425\linewidth]{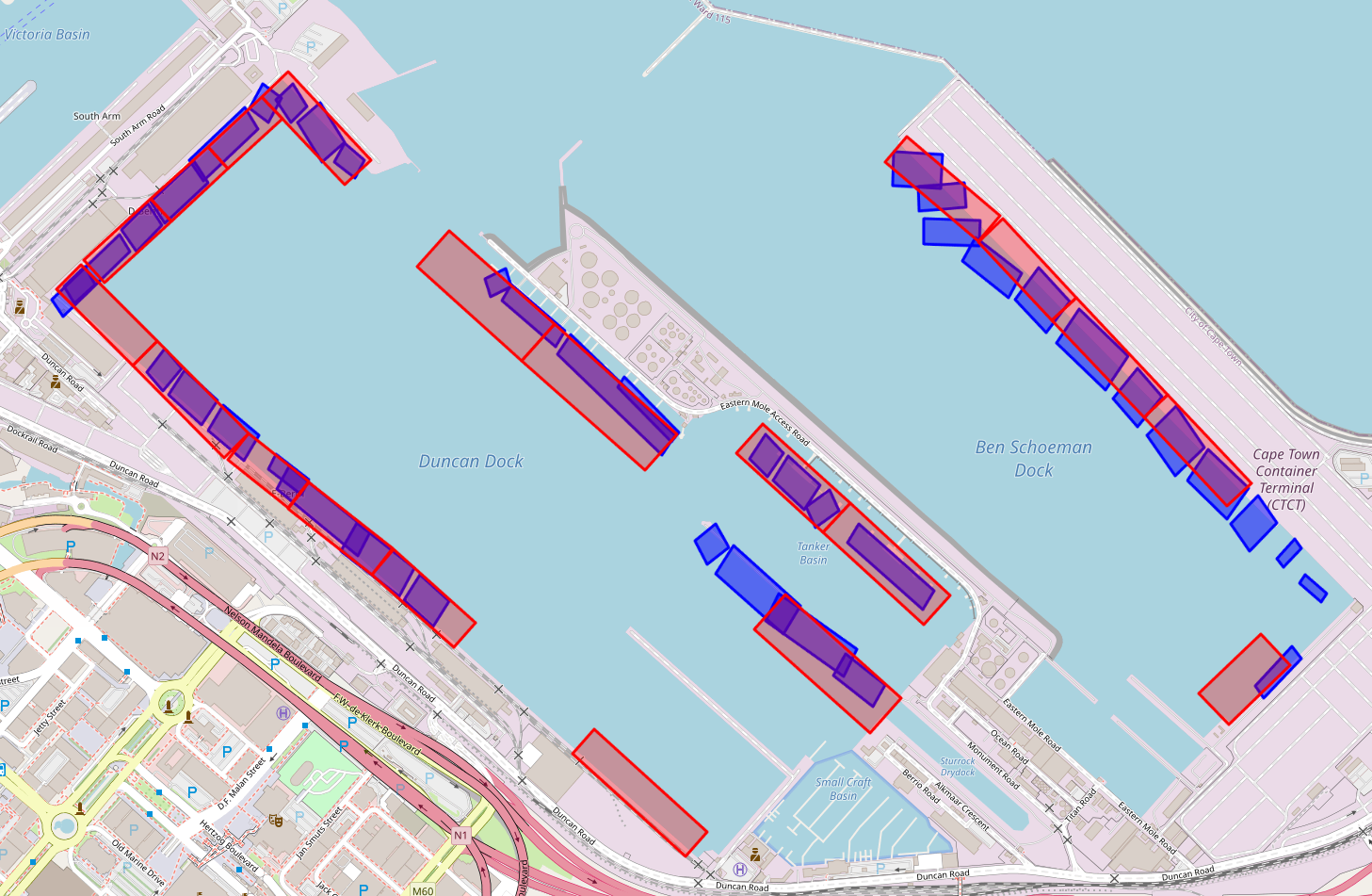}\\
    \includegraphics[width=0.425\linewidth]{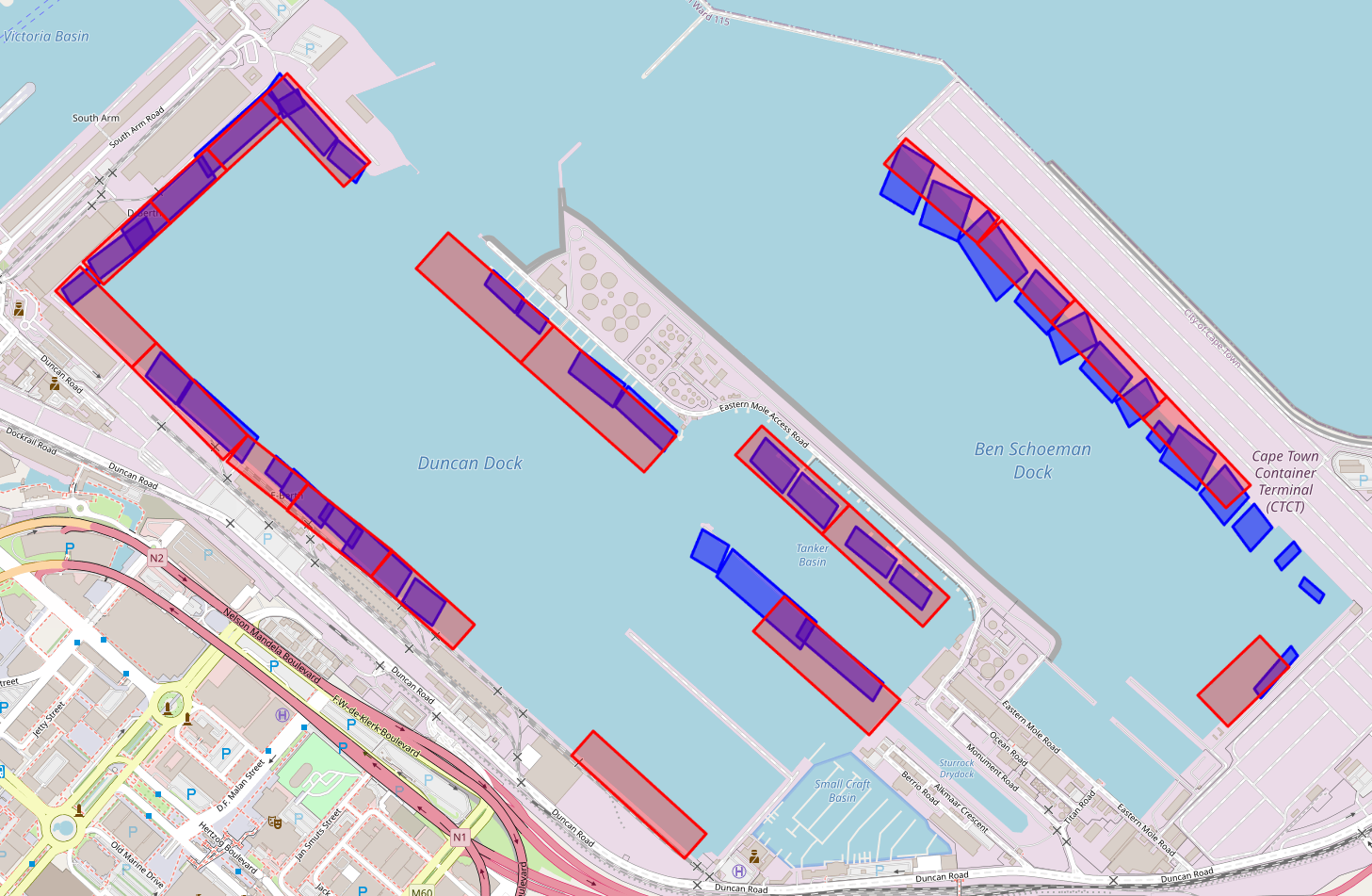}
    \includegraphics[width=0.425\linewidth]{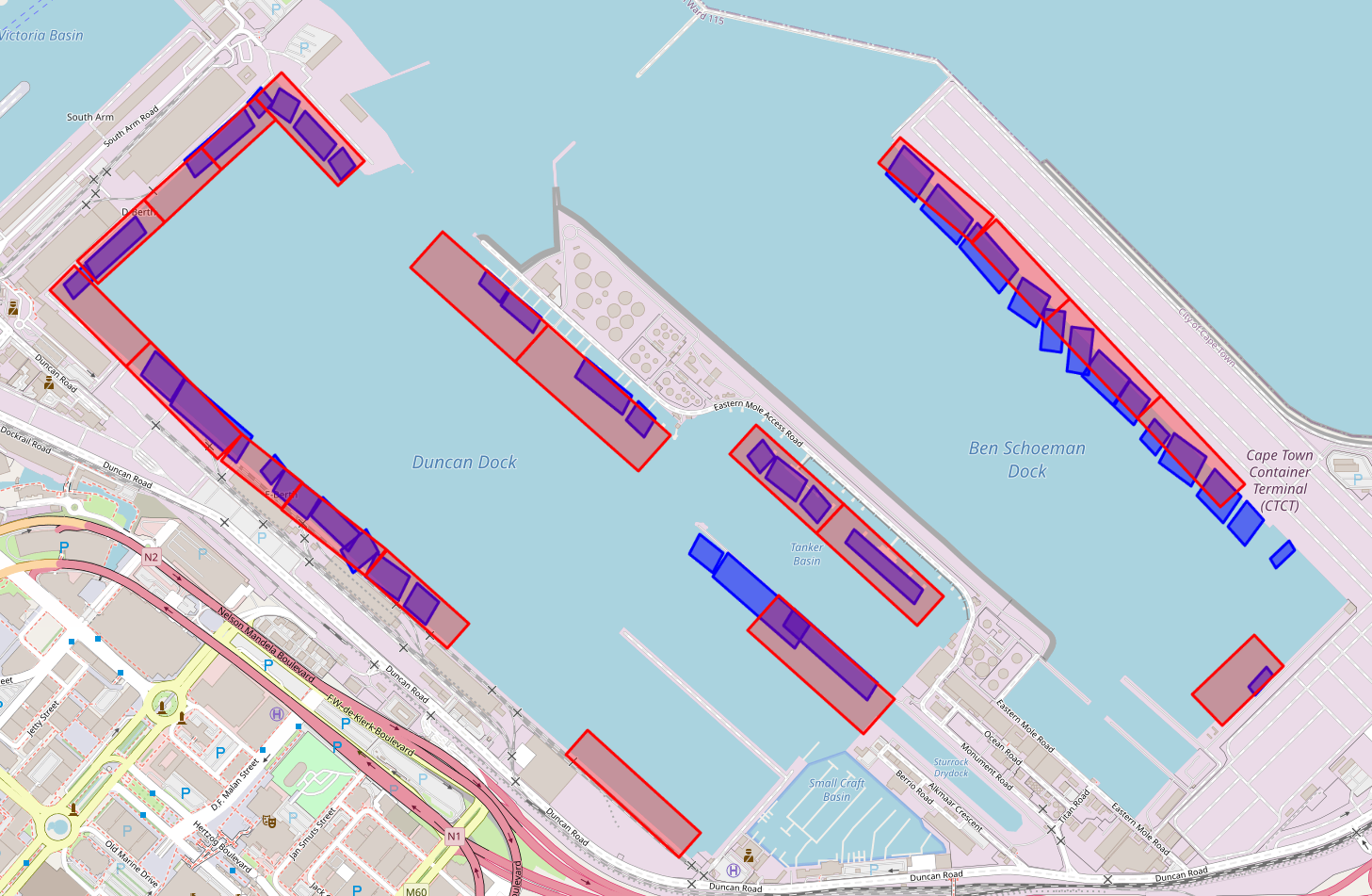}
    % \\
    % \includegraphics[width=0.49\textwidth]{geo_15T.png}
    % \includegraphics[width=0.49\textwidth]{geo_30T.png}
    % \\
    % \includegraphics[width=0.49\textwidth]{geo_1h.png}
    % \includegraphics[width=0.49\textwidth]{geo_2h.png}
    
% \end{adjustwidth}
\caption{Qualitative results for the ablation study of interpolation periods for the port of Cape Town using the non-geohash variant. From left to right and top to bottom the interpolation period is 15 minutes,  30 minutes, 1 hour, 2 hours.}
\label{fig:inter_ablation_qual}
\end{figure}  

% This ablation study evaluates the sensitivity of the non-geohash variant to different interpolation periods. 
As shown in Table~\ref{tab:ablation_gen}, shorter intervals (notably 15 minutes) again achieve slightly better quantitative results, but at the cost of increased computational complexity. Unlike the geohash variant, which exhibited unnatural berth delineations at very short intervals, the non-geohash variant does not present such clear qualitative drawbacks (see Figure~\ref{fig:inter_ablation_qual}), making it difficult to identify a definitive winner.Nevertheless, these differences are not substantial enough to justify adopting a separate configuration from the geohash-based method. Given the acceptable performance at 1 hour and the associated efficiency benefits, we maintain the same 1-hour interpolation period for both variants. This ensures consistency across experiments while balancing performance and computational cost.
\bibliographystyle{unsrt} 
\footnotesize
\bibliography{bib}

% %% For authoryear reference style

% \bibitem[Lamport(1994)]{lamport94}
%   \textit{\LaTeX: a document preparation system},
%   Addison Wesley, Massachusetts,

% \end{thebibliography}
\end{document}